\documentclass{article}

    \PassOptionsToPackage{numbers, compress}{natbib}

    \usepackage[preprint]{neurips_2021}

\usepackage[utf8]{inputenc} 
\usepackage[T1]{fontenc}    
\usepackage{hyperref}       
\hypersetup{colorlinks=true}
\usepackage{url}            
\usepackage{booktabs}       
\usepackage{amsfonts}       
\usepackage{nicefrac}       
\usepackage{microtype}      

\usepackage{graphicx}
\usepackage{amsbsy}
\usepackage{color}

\newcommand{\gauss}{\mathcal{N}}

\newcommand{\db}{{\bf d}}

\title{
Sifting out the features by pruning:\\ Are convolutional networks the winning lottery ticket of fully connected ones? }

\author{ 
  Franco Pellegrini\\
  Laboratoire de Physique de l'\'{E}cole normale sup\'{e}rieure, ENS, Universit\'{e} PSL, \\
  CNRS, Sorbonne Universit\'{e}, Universit\'{e} de Paris ---
  F-75005 Paris, France\\
  \texttt{franco.pellegrini@phys.ens.fr}
  \And
  Giulio Biroli\\
  Laboratoire de Physique de l'\'{E}cole normale sup\'{e}rieure, ENS, Universit\'{e} PSL, \\
  CNRS, Sorbonne Universit\'{e}, Universit\'{e} de Paris ---
  F-75005 Paris, France\\
  \texttt{giulio.biroli@phys.ens.fr }
}

\begin{document}
\maketitle
\begin{abstract}
Pruning methods can considerably reduce the size of artificial neural networks without harming their performance. In some cases, they can even uncover sub-networks that, when trained in isolation, match or surpass the test accuracy of their dense counterparts.
Here we study the inductive bias that pruning imprints in such ``winning lottery tickets''.
Focusing on visual tasks, we analyze the architecture resulting from iterative magnitude pruning of a simple fully connected network (FCN).
We show that the surviving node connectivity is local in input space, and organized in patterns reminiscent of the ones found in convolutional networks (CNN).
We investigate the role played by data and tasks in shaping the architecture of pruned sub-networks.
Our results show that the winning lottery tickets of FCNs display the key features of CNNs.
The ability of such automatic network-simplifying procedure to recover the key features ``hand-crafted'' in the design of CNNs suggests interesting applications to other datasets and tasks, in order to discover new and efficient architectural inductive biases.
\end{abstract}

\section{Introduction}

In the last decade deep and over-parametrized neural networks achieved breakthroughs in a variety of contexts~\cite{Alex12,He16,Vaswani17,GPT3,Silver17}. To keep pushing the boundaries of their capacity, these networks have grown in size to over billions of parameters~\cite{Tan19}. Network pruning approaches have recently received renewed attention as effective procedures to counterbalance this growth by reducing networks size without harming their performance. Several pruning methods have been introduced~\cite{Hoefler21}: they are all based on the idea of cutting weights from a trained, or partially trained, network in order to obtain a sparse sub-network that works at a comparable level of accuracy than the original one. For applications, pruning is very efficient in improving storage and computational cost. On the theoretical side, it led to the discovery that within a randomly-initialized network one can find a sub-network, {\it a winning lottery ticket}, that when trained in isolation can match the test accuracy of the original network after training for at most the same number of iterations~\cite{Frankle19}.

The introduction of better network architectures, with geometries more suited to specific tasks, and which result in faster training and overall better generalizing properties can also be seen from a similar perspective. For instance, Convolutional Neural Networks~\cite{Alex12,He16,LeCun89,Simonyan15,Szegedy15} (CNN), which are very efficient for visual tasks, can be embedded in the Fully Connected Network (FCN) class. CNNs can therefore be interpreted  as sparse and smaller FCNs trained from very specific initial conditions very well adapted to the task~\cite{Dascoli19}. Also in this case the size is drastically reduced (with respect to the FCN counterpart). The difference with pruning is that the main characteristics of CNNs---local connectivity and weight sharing---do not come from an automatic procedure but are ``hand-designed'' features introduced starting from analogies with the virtual cortex~\cite{Hubel62,Fukushima80}, studies of the invariant properties of natural images~\cite{Ruderman94}, and engineering improvements~\cite{LeCun89}.
Pruning instead proceeds in an agnostic way in order to find smaller and efficient architectures; it takes advantage of properties which are imprinted in the network through training and which result from the interplay of data, task and optimization algorithm.
Pruning induces an {\it inductive bias} customized to the learning task at hand, which leads to a network with better training and generalization properties compared to one of similar size trained from scratch.  
It has been actually shown that the same winning lottery ticket generalizes across training conditions and similar datasets~\cite{Morcos19}. This implies that the bias induced by pruning is sufficiently generic to transfer the pruned networks within the same data domain. 

Here, we characterize the nature of such inductive bias in the case of dense networks trained to classify natural images, and reveal that the winning lottery tickets of FCNs display the key features of CNNs. We apply \textit{iterative magnitude pruning} (IMP)~\cite{Frankle19} on FCNs trained on a low resolution version of ImageNet~\cite{ImageNet32}.
We show that the sub-network obtained by IMP is characterized by {\it local connectivity}, especially in the first hidden layer, and masks leading to {\it local features} with patterns very reminiscent of the ones of trained CNNs~\cite{Zeiler14}. Deeper layers are made up of these local features with larger receptive fields hinting at the hierarchical structure found in CNNs.
We study how the data and the task affect these properties. We show a crossover from a large dataset regime (large signal-to-noise ratio), where the inductive CNN-like bias is present, to a small dataset regime in which pruning loses performance and concomitantly the properties described above disappear. 
We show that a similar crossover takes place when going from a meaningful task to one with low semantic value.

\section{Related works} 
Several recent works have studied pruning from the perspective of the lottery ticket hypothesis~\cite{Frankle19}: 
some investigated more general weight reinitializations~\cite{Zhou19}, whereas others actually questioned its validity on larger nets~\cite{Liu19,Gale19}.  Later, Refs.~\cite{Frankle19-2,Frankle20} showed that in order to obtain good results for complex networks, datasets, and tasks it is important to modify the rewinding time compared to the initial formulation of the hypothesis. Other studies concerned hyperparameters modifications~\cite{Frankle20-2,Renda20}, and concentrated on the transferability~\cite{Morcos19,Sabatelli20} of the pruned networks.
To the best of our knowledge, our work is the first to analyze the internal structure of winning tickets in order to highlight the inductive bias induced by pruning and relate it to architectural properties.  

Perhaps the most related work is Ref.~\cite{Neyshabur20}, which addresses the issue of learning CNN-like inductive bias from data and through training. It shows that training using a modified $\ell_1$ regularization is a way to induce local masks for visual tasks. This is similar to our results on pruning: enforcing sparsity during training allows to recover locality.
Ref.~\cite{Dascoli19} studies the role of CNN-like inductive biases by embedding convolutional architectures within the general FCN class. It shows that enforcing CNN-like features in an FCN can improve performance even beyond that of its CNN counterpart.  
Finally, a manuscript~\cite{tolstikhin2021mlp} that appeared after the completion of our work shows that by considering a particular multilayer perceptron architecture, called MLP-mixer, 
some of the CNN features can be learned from scratch using a large training dataset. 

%
\section{Method, observables and notation}\label{ss:setup}
\paragraph{Data, networks, and training procedure} Throughout this work our reference dataset, referred to as ImageNet32, consists of the almost 1.3M images classified in 1000 categories of the ILSVRC-12 image classification challenge~\cite{ILSVRC15}, cropped and scaled down to $32\times32$ resolution as detailed in~\cite{ImageNet32}. Considering the 3 color channels (RGB) the input size is $n_0=3072$ (we index the input as layer 0). The validation test contains 10k images.
To keep the analysis simple, in the following we consider a FCN having 3 hidden layers of equal size $n_1=n_2=n_3=1024$. Generalizations to more general FCN architectures (more layers and different layer sizes) are presented in the SM \ref{SMss:arch}.
Note that the aim of our work is to characterize the sub-networks induced by pruning in the simplest setup which allows a thorough analysis. For this reason, we use a   
small and fixed input size, which is useful to have a manageable size for the FCN, and we do not use any form of augmentation in order to avoid biases towards translational invariant features (the choice of a large dataset is important in this sense to obtain good statistics, as shown in Sec.~\ref{s:roledata}).
We apply batch normalization~\cite{Ioffe15} and we use rectified linear unit (ReLU) nonlinearities for each layer.
We denote $w^{l+1}_{ij}$ the weights between node $i$ of layer $l$ and node $j$ of layer $l+1$. Weights are initialized from a normal distribution $w^{l+1}\sim\gauss(0,2/(n_l+n_{l+1}))$~\cite{Glorot10}. Each node also has a bias $b^l_i$, initialized at 0.
The size of the output layer equals the number of categories ($n_4=1000$ in this first experiment, 10 in Section~\ref{s:roletask}), which are converted to probabilities to compute the cross-entropy loss. The reported accuracy is the ratio of correct, most probable, labels.
We train on mini-batches of 1000 images, and minimize through stochastic gradient descent with learning rate $\alpha=0.1$. Each training run corresponds to $10^5$ steps (around 78 epochs): while this is not sufficient to fit the training set, it is effective in identifying a winning lottery ticket~\cite{You20}. 
We checked that our results are robust while changing the optimization procedure, see the SM \ref{SMss:min}.

\paragraph{Iterative Magnitude Pruning} We prune iteratively the 3 hidden layers. Each time, we remove $p=30\%$ of previously unpruned weights with the lowest absolute value at the end of training, per layer, as originally described in~\cite{Frankle19}.
When analyzing the pruned networks, we denote $u$ the ratio of unpruned weights per layer: after $n$ iterations $u=(1-p)^n$.
As suggested in~\cite{Frankle19-2}, we rewind the weights after each training not to the initial values, but to the ones obtained after 1000 steps of optimization in the first run. Instead of a winning lottery ticket the resulting network has been called a matching ticket~\cite{Frankle19-2} (we have verified that the main result holds when rewinding to 0, see SM \ref{SMss:IMPpar}). We iterate the training and pruning until less than 20\% of the nodes in any layer retain any connection to the previous layer, i.e.\ until more than 80\% of the nodes have been completely pruned away.
More information on the pruning parameters, including total training time, is presented in SM \ref{SMss:IMPpar}.

\paragraph{Observables} In order to analyze the structure of the subnetwork induced by pruning we focus on the following observables:
\begin{itemize} 
\item \textit{Masks}: for each layer, we define masks by assigning 0 to pruned weights, and 1 to all the others. We indicate with $m^l_{ij}$ the mask formed by the unpruned weights $w^l_{ij}$.
\item{\it Node connectivity}: for each node $j$ in layer $l$ we define the input connectivity $C^{{\rm in},l}_j=\sum_i m^l_{ij}$ as the number of input unpruned nodes, and $C^{{\rm out},l}_j=\sum_i m^{l+1}_{ji}$ the output connectivity. 
\item {\it Local distances}: in order to asses the locality of the masks, we count the number of pixels in a given relative position, $\db$, that are connected to the same hidden node, as shown schematically in Fig.~\ref{fig:2}\textbf{a}. 
More precisely: two inputs $i$ and $i'$ are connected through node $j$ if $m^1_{ij}m^1_{i'j}=1$. 
For such inputs we define the displacement $\db_{ii'}$: if input index $i$ identifies a pixel of spatial position $(x,y)$ and $i'$ corresponds to position $(x',y')$, we define their displacement $\db_{ii'}\equiv(x'-x,y'-y)$. 
The histogram of connected pixels at a given displacement is denoted  $S(\db)=\sum_j\sum_{i,i'|\db_{ii'}=\db}m^1_{ij}m^1_{i'j}$. 
We consider both the sum over pixels belonging to the same color channel ($S^{\rm sc}$) and over different color channels ($S^{\rm dc}$).
\end{itemize}

%

\section{Pruning sifts out local features and recovers CNN-like masks}\label{s:main}
\begin{figure}[!ht]
  \centering
  \includegraphics[width=\columnwidth]{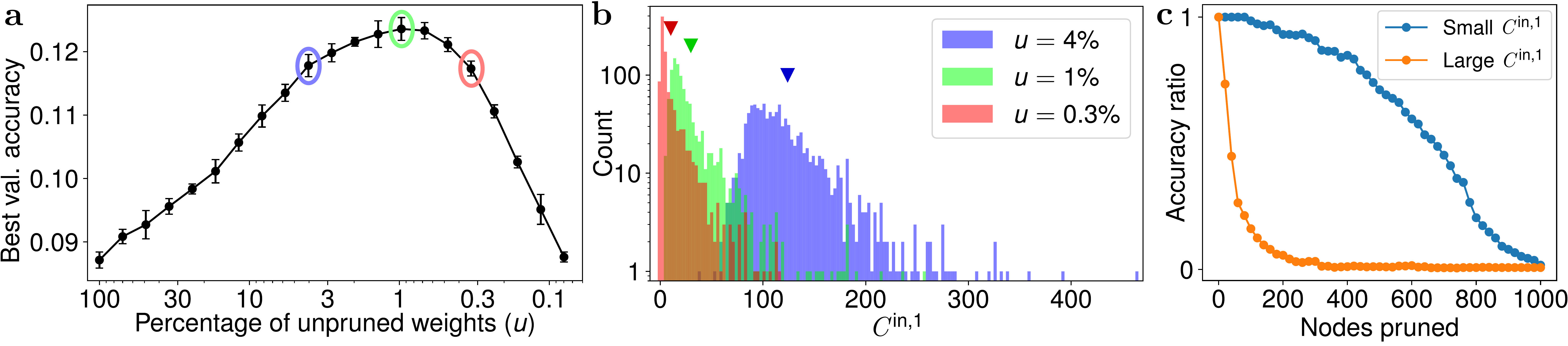}
  \caption{
    \textbf{a:} Highest validation accuracy at different percentage of unpruned weights. Averages and standard deviations over 4 independent IMP runs.
    \textbf{b:} Histograms of the number of $C^{{\rm in},1}$ for three configurations at different sparsities. The triangles represent the average connections $n_0u$ at each sparsity.
    \textbf{c:} Accuracy with respect to the full network for networks with a varying number of nodes removed, in increasing or decreasing order of $C^{{\rm in},1}_j$.
  }\label{fig:1}
\end{figure}
\begin{figure}[!ht]
  \centering
  \includegraphics[width=\columnwidth]{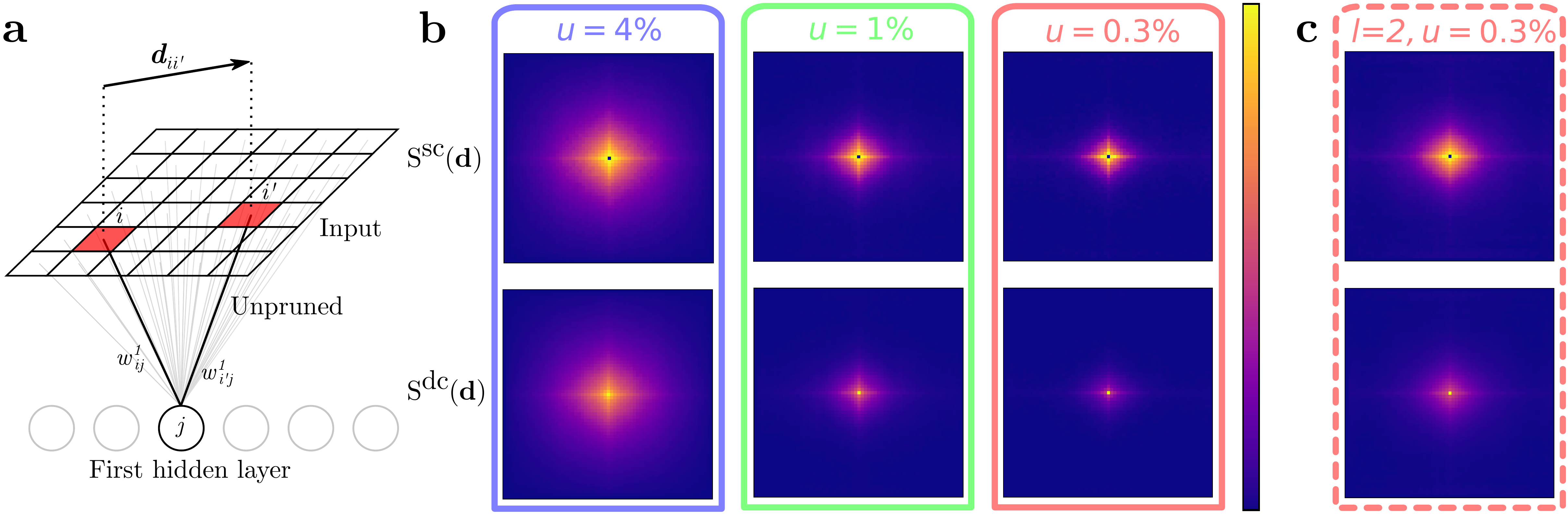}
  \caption{
    \textbf{a:} Sketch of the displacement and connectivity considered when computing $S(\db)$.
    \textbf{b:} $S^{\rm sc}$ (top) and $S^{\rm dc}$ (bottom) for each sparsity considered (columns).
    Relative positions are shown with respect to the center of each image ($\db=0$) and the color scale of each plot goes from 0 to its maximum. For $S^{\rm sc}$, the single point at $\db=0$ is removed as it is trivially always connected.
    \textbf{c:} Same as \textbf{b}, but for the second hidden layer at the higher sparsity.
  }\label{fig:2}
\end{figure}

Fig.~\ref{fig:1}\textbf{a} shows the highest (early stop) validation accuracy as a function of $u$ throughout the IMP procedure. As expected~\cite{blalock2020state}, pruning allows to find sub-networks of considerably smaller sizes that work as well as the original ones. Actually, as shown in Fig.~\ref{fig:1}\textbf{a}, IMP leads to a strong improvement in the accuracy which goes from less than 9\% to more than 12\% when 99\% of the weights are pruned\footnote{The overall accuracy is low compared to the state of the art. This is due to the fact that we are focusing on a simple (hence fully analyzable) setup and by choice we avoid improvements that could lead to biases in the results. Yet, considering the low resolution, lack of augmentation, and simplicity of the network for a notoriously complex dataset, the improvement over the random $10^{-3}$ is a clear signal of learning. The top-5 accuracy is around $26\%$.}.

\paragraph{The pruning-induced masks are local} To characterize these pruned network, we concentrate on 3 sparsity values close to the maximum accuracy reached by IMP: $u=4\%$, 1\%, and 0.3\%, as circled in Fig.~\ref{fig:1}\textbf{a}.
We start by showing the histograms of $C^{{\rm in},1}$, i.e.\ the input connectivity per node of the first hidden layer. For a random pruning, this would be binomially distributed around the value $n_0u$, i.e.\ 123, 31, and 10 respectively for the sparsities considered.
Fig.~\ref{fig:1}\textbf{b} shows that the distributions obtained after pruning instead have long positive tails, representing a population of nodes that preserve a large number of unpruned input weights.
Since IMP should preserve ``important'' connections, it is interesting to check whether these nodes represent meaningful \textit{features} that have been sifted out by the procedure.
To this end, we consider the relative accuracy of the final network when either the nodes with the largest or with the smallest $C^{{\rm in},1}$ have been completely removed (i.e.\ we set $m^1_{ik}=0$ for those nodes $k$, without any further training).
As shown in Fig.~\ref{fig:1}\textbf{c}, the accuracy as a function of number of removed weights drops much more quickly when removing the highly connected nodes, thus confirming their relevance. 

In order to investigate the emerging local structure of the unpruned weights, we consider the number of connected pixels in a given relative position $S(\db)$.
A 2D map for any possible displacement $\db$ is shown in Fig.~\ref{fig:2}\textbf{b}, for the 3 sparsities considered and for either same color or different color channels. Remarkably, we find that inputs connected to the same node are locally close, i.e. their distance is small, both within and between color channels. Note that the same 2D maps for the unpruned layer or pruned layers with random connections, are very different and non-local as shown in the SM \ref{SMs:masks}.
We also find that straight horizontal and vertical relative displacements are slightly more likely.
As the pruning progresses, the localization becomes stronger and the anisotropy more apparent. The same 2D maps are shown by only selecting inputs connected to nodes with a given connectivity in the SM \ref{SMss:locconn}.
Our results show that the locality of the masks holds for all nodes except the ones with the smallest values of $C^{{\rm in},1}$. For the larger masks the effect is truly remarkable given their large number of unpruned weights. The set of results discussed above unveils one key feature of the matching (and winning) lottery ticket of FCNs: they display {\it local masks} that emerge by pruning.   
\begin{figure}[!hbt]
  \centering
  \includegraphics[width=\columnwidth]{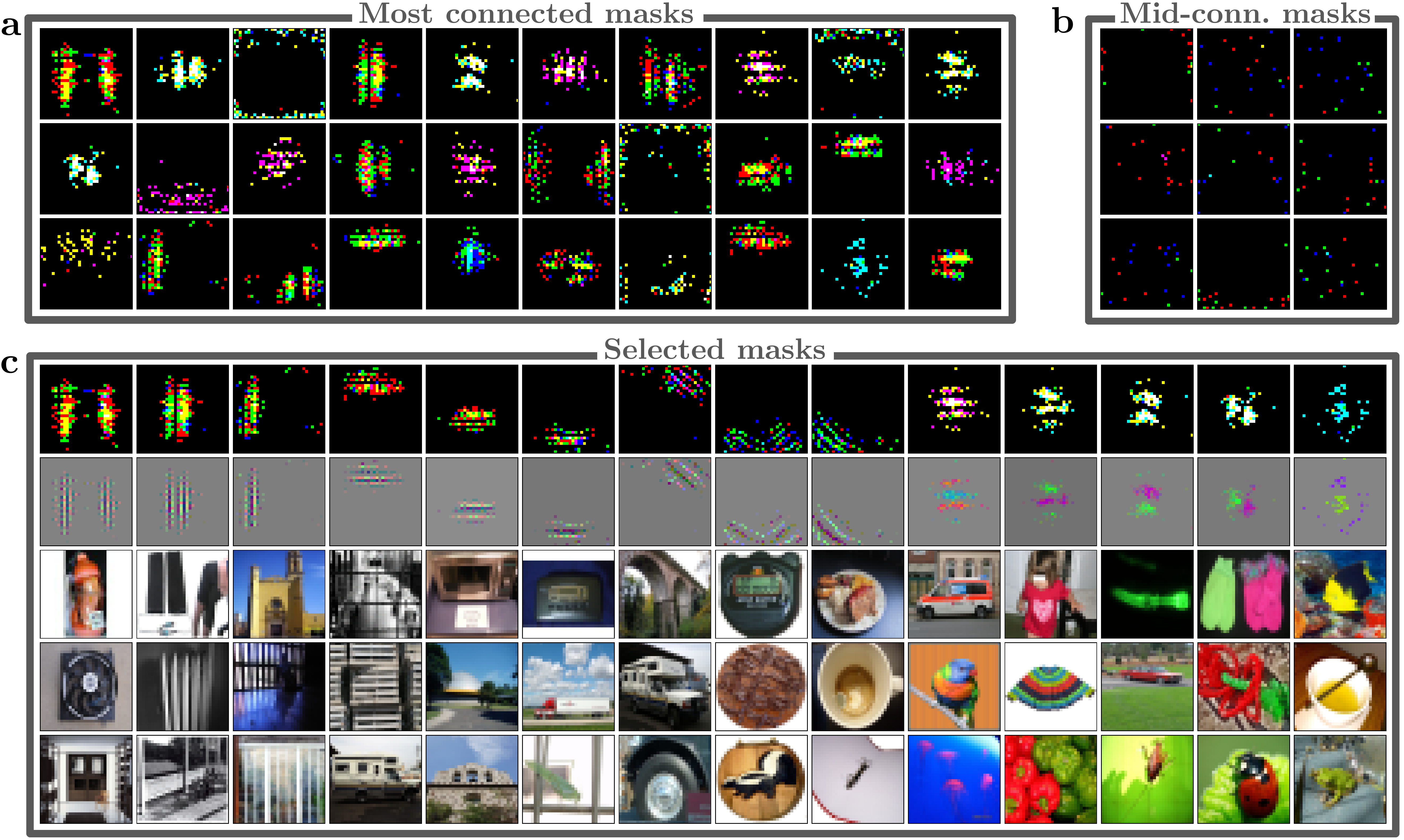}
  \caption{
    \textbf{a:} Masks of the 30 nodes with largest $C^{{\rm in},1}_j$ for the best IMP iteration. Binary masks are directly converted to RGB intensities.
    \textbf{b:} Masks of 9 sample nodes around half in the $C^{{\rm in},1}_j$ ranking (having 20 connections each). Same iteration and representation as panel \textbf{a}.
    \textbf{c:} For a hand-picked selection of the most connected nodes (columns), the top row reports the mask and the second row represent the masked weights $w^1_{ij}m^1_{ij}$ normalized from minimum to maximum and converted to RGB. The last 3 rows represent the images with highest activation for that specific node.
  }\label{fig:3}
\end{figure}

\paragraph{The local features are CNN-like}
We now concentrate on the nodes with the largest $C^{{\rm in},1}_j$, which have been found to be most relevant, and focus on  the spatial structure of the associated masks. 
Fig.~\ref{fig:3}\textbf{a} shows the masks of the 30 most connected nodes for the best IMP iteration ($u\simeq 1\%$): they all present well localized distributions, either in a single location, symmetric with respect to the vertical axis, or sometimes around the edges of the images.
Several masks also retain the same pixels through multiple color channels, often with a well defined pattern (e.g.\ 2 channels out of 3).
Moreover, many of the masks show a specific preference for vertical and horizontal directions, in line with the previously observed anisotropy, with some consisting of parallel lines separated by one or two pixels (the effect is even more pronounced for stronger pruning, see SM \ref{SMss:bestm}, where we also show a larger sample of masks). A video of the evolution of first layer masks is linked in SM \ref{SMs:ext}.
Masks with a lower number of connections, e.g.\ the ones positioned around the half of the $C^{{\rm in},1}_j$ ranking, do not show such well defined structures, as shown in Fig.~\ref{fig:3}\textbf{b}.
In Fig.~\ref{fig:3}\textbf{c} we select a few nodes within the top 5\% most connected ones to highlight their specific structure: vertical lines, horizontal lines, oblique lines and complementary patches of color.
To clarify the features selected by these nodes, we show not only the masks, but also the masks multiplied by the final weights and the 3 images in the validation with the highest activations over these nodes. 
The weights highlight further structure in the localized patches, with alternating positive and negative weights perpendicular to the preferential direction for long masks, or between different patches.
These are very reminiscent of Gabor filters~\cite{Marcelja80}, and very similar to CNN masks, which respond preferentially to local high contrast details, e.g.\ edges in a specific direction or color patches~\cite{Zeiler14}. 
Moreover, very comparable filters can be seen at different locations, showing the emergence of partial translational invariance in the local features even without data augmentation\footnote{Note that images in the training set have a bias for structure in the center and at the boundary, a characteristic that emerges also from pruning: features tend to have more connections towards the center and edges of the image, as shown in SM \ref{SMs:masks}.}.
Lastly, the example images unveil the response of these nodes also to local features not directly related to the final classification task. This directly relates to the good transfer properties of the subnetworks corresponding to the winning tickets \cite{Morcos19, Sabatelli20}. 
Further experiments highlighting the structure of the masks are presented in SM \ref{SMss:rot}, where the images are rotated, in SM \ref{SMss:IN64}, where higher resolution images are used as input, and in SM \ref{SMss:transl}, where we restore translational invariance by considering all possible image translations.

To continue the parallel with CNNs, we now focus on the second hidden layer. In this case, for a given second layer node $j$, we define the associated effective mask as $\mu_{ij}=\theta(\sum_k m^1_{ik}m^2_{kj}-1/2)$ ($\theta$ being the step function), which equals 1 only if pixel $i$ is connected to \textit{any} first hidden layer node connected to $j$, and 0 otherwise.
As done for the first hidden layer, we asses the locality of these masks via the local displacements, $\db_{i,i'}$, of the input nodes connected to the same mask. We show the 2D map of these displacements, $S(\db)$, in Fig.~\ref{fig:2}\textbf{c} for same and different channels for pruning $u=0.3\%$. 
Since the network is just 3 layers deep, and the receptive fields are already fairly large in the first layer, one does not expect the second layer nodes to be extremely localized. Yet, we find once again fairly local masks which favor horizontal and vertical directions, thus showing that pruning induces locality also beyond the first hidden layer. Masks from this layer, and some details on their composition, are further analyzed in SM \ref{SMss:bestm}.

\section{The number of data controls pruning performance}\label{s:roledata}
\begin{figure}[!htb]
  \centering
  \includegraphics[width=\columnwidth]{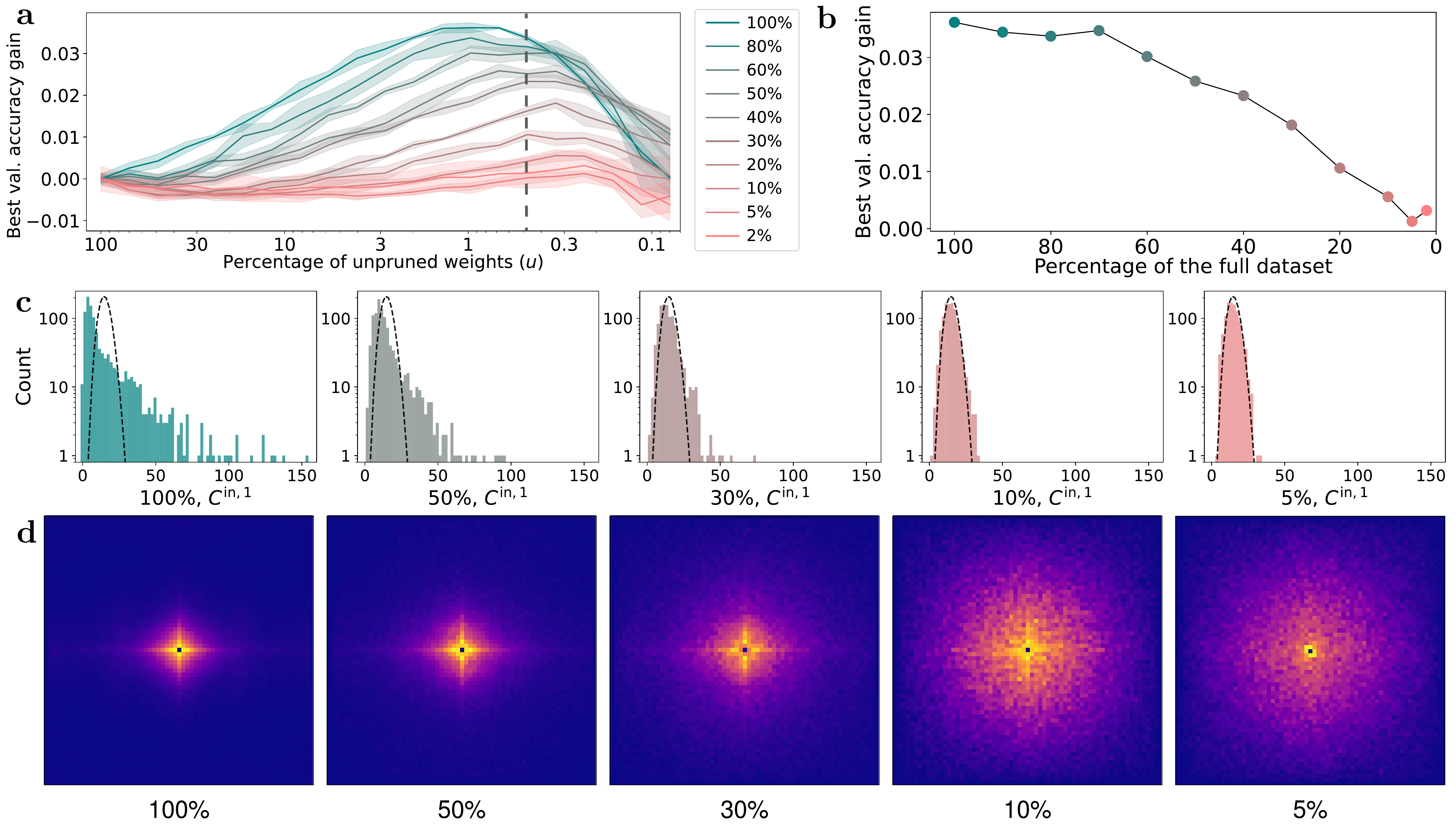}
  \caption{
    \textbf{a:} Gain in best validation accuracy (difference from the value corresponding to the unpruned network) as a function of $u$ for training on different percentages of the original dataset. 
    The lines are averages of 4 independent runs, with the shaded area representing one standard deviation.
    The dashed line marks the iteration at $u\simeq0.5\%$ used in panels \textbf{c} and \textbf{d}.
    \textbf{b:} Largest gain in validation accuracy during the IMP procedure for different dataset sizes (percentage of the whole dataset).
    \textbf{c:} Histogram of $C^{{\rm in},1}$ for different dataset sizes (see label at the bottom) at the iteration $u\simeq0.5\%$. The dashed lines represent the theoretical binomial distribution for random pruning.
    \textbf{d:} $S^{\rm sc}(\db)$ for different dataset sizes at the iteration $u\simeq0.5\%$. Normalization and color map as show in Fig.~\ref{fig:2}.
  }\label{fig:4}
\end{figure}
We now study to what extent the effects found in the previous section depend on the properties of the dataset, in particular in the following we focus on the role of {\it the number of data} used to train the network. 
We repeat the training and the IMP with datasets of progressively smaller size, down to 2\% of the original. While this may seem like an extreme reduction of the dataset, even just 5\% of the data is actually larger than commonly used datasets such as CIFAR-10/100.
What interests us here is the gain in accuracy due to the IMP, i.e. 
how much pruning leads to an increase of performance for a given number of data\footnote{Of course, the best validation accuracy decreases as we decrease the dataset size (to just around 2\% for 5\% of the data).}. 
In Fig.~\ref{fig:4}\textbf{a} we show the difference in accuracy with respect to the value obtained for the unpruned network during the IMP procedure. When decreasing the size of the dataset, 
the benefit of pruning diminishes, until no improvement at all is visible for the smallest datasets, see panel \textbf{b}, showing the maxima as a function of the dataset size.
Decreasing the number of data used to train the network also drastically affects the connectivity and locality of the subnetworks found by pruning. 
In order to present this result we focus for each dataset size on an iteration close to the maximum of all curves of panel \textbf{a}, around $u=0.5\%$. 
Panels \textbf{c} show that decreasing the dataset size the connectivity distribution becomes essentially identical to the one corresponding to pruning weights at random independently of their magnitude (dashed line). Concomitantly, masks become more and more non-local, as shown in panels \textbf{d} (note that low connected nodes are affected earlier, see SM \ref{SMss:locconn}).

Our results show that the dataset size clearly plays a major role for pruning. Our interpretation is that by decreasing the number of data one decreases the signal to noise ratio, hence effectively increasing random idiosyncratic fluctuations in the weights. When such fluctuations become of the same order of magnitude as the ones of the good subnetworks embedded in the original dense one, 
it becomes impossible to unveil the matching lottery ticket by pruning. In consequence, in this regime the subnetworks obtained by IMP are random, featureless and do not display any local feature.  
More results on the disruption of local features due to small dataset sizes are shown in SM \ref{SMss:bestm}.

\section{The role of the task: the importance of being meaningful}\label{s:roletask}
\begin{figure}[!htb]
  \centering
  \includegraphics[width=\columnwidth]{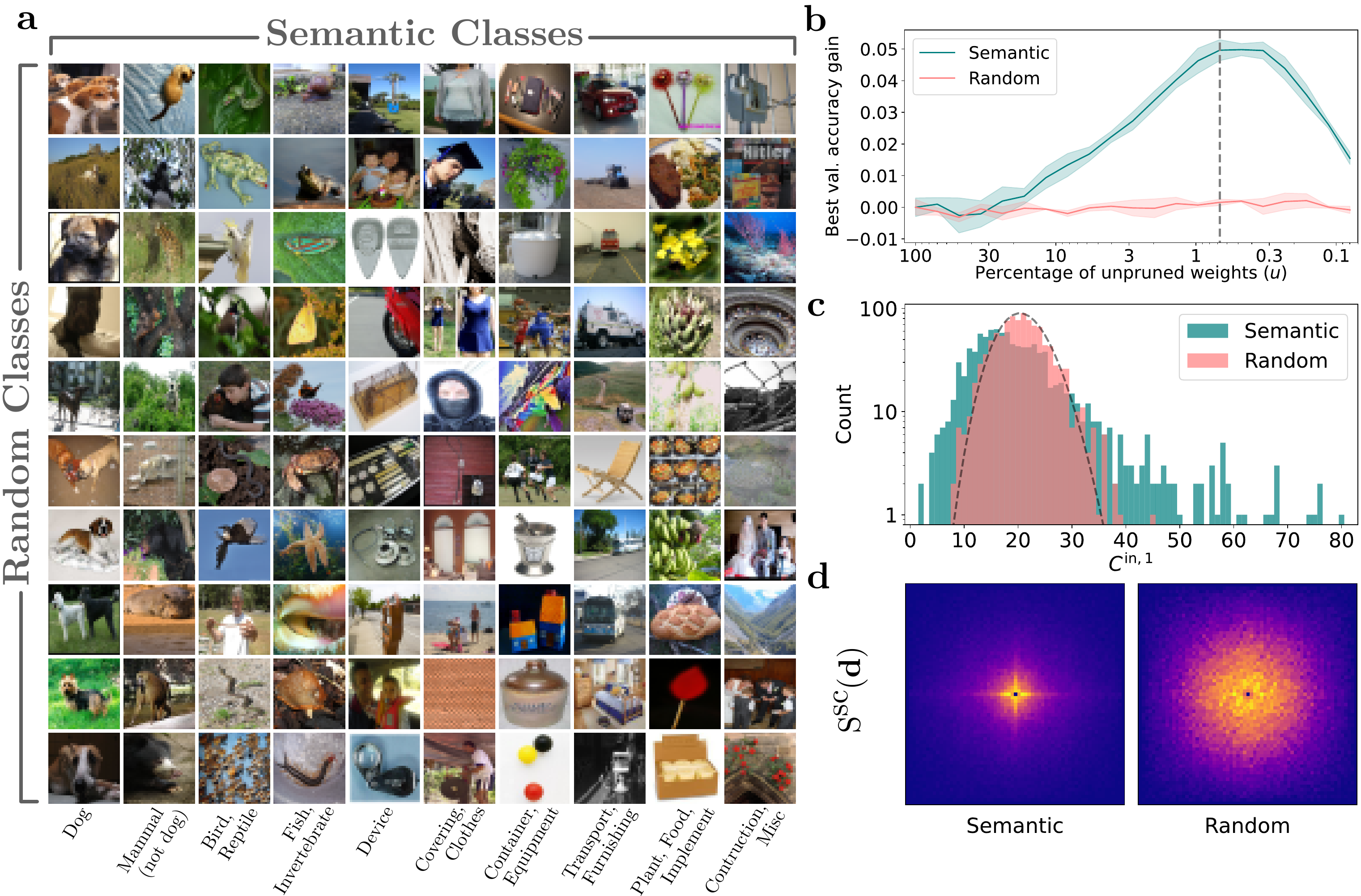}
  \caption{
    \textbf{a:} 10 sample images for each of the 10 macro classes. Each column shows a different \textit{semantic} cluster (rough label at the bottom) where images can be seen to have similar content, while each row is a different \textit{random} cluster, and no structure is visible.
    \textbf{b:} Gain in best validation accuracy (difference from the value corresponding to the unpruned network) for the random and semantic tasks. 
    The lines are averages of 4 independent runs, with the shaded area representing one standard deviation.
    Each training was run for 500k steps for the semantic and 300k steps for the random clustering to ensure convergence.
    The dashed line marks the iteration at $u\simeq0.7\%$ used in panels \textbf{c} and \textbf{d}.
    \textbf{c:} Histogram of $C^{{\rm in},1}$ for the two cases at the iteration $u\simeq0.7\%$. The dashed line is the theoretical binomial distribution for random pruning.
    \textbf{d:} $S^{\rm sc}(\db)$ for the two cases at the iteration $u\simeq0.7\%$. Normalization and color map as shown in Fig.~\ref{fig:2}.
  }\label{fig:5}
\end{figure}
To further understand the role played by the properties of the dataset we now modify it in the following way. We cluster the original 1000 classes of the training set and test set in 10 balanced macro-classes in two different ways: the first one, called \textit{random}, groups the original categories in a non-meaningful way (on the basis of their numerical label in the dataset modulo 10). This produces 10 classes which are all very similar. The second one, called \textit{semantic}, clusters together similar original classes, into macro-categories such as ``dog'', ``device'', ``transport and furnishing'', etc. 
This produces 10 classes which characterize the data in a meaningful way. An example of the difference between these two ways of regrouping the 1000 classes is reported in Fig.~\ref{fig:5}\textbf{a}, where the rows represent examples from the ``random'' classes, and the columns from the ``semantic'' classes. For more details, see SM \ref{SMs:clust}.
After training the accuracy obtained for the random task is at best 13\%, while it surpasses 36\% for the semantic one. This difference is expected since in the former case elements of different classes are very difficult to be distinguished and training mainly leads  to memorization and not learning. 

We now compare the gain in accuracy due to pruning for the semantic and the random task. Fig.~\ref{fig:5}\textbf{b} shows that pruning is beneficial for the former but not for the latter. Moreover, the geometrical properties of the subnetwork induced by pruning are very different in the two cases. In panel \textbf{c} and panel \textbf{d} we show respectively the histogram of 
the connectivities $C^{{\rm in},1}$ and of the local displacements $\db_{ii'}$. These results 
reveal that the subnetwork obtained by pruning 
displays local features and non-trivial connectivity distribution for the semantic task only. In the other case, the subnetwork is essentially featureless, non-local and like one where weights have been pruned at random. Sample masks for the two tasks are shown in SM \ref{SMs:masks}.

These results highlight the role of the task in shaping the properties of the network obtained by pruning: only for the task that the network can efficiently learn, and not just memorize, local features emerge. 
This result is a direct consequence of the \textit{qualitative} difference in the network solutions for the two tasks: features do not simply emerge from the data in an unsupervised fashion, but they represent structures useful for optimization with respect to a meaningful loss. If the task is solved in an unstructured way, e.g.\ through memorization, the features are simply not selected.
In SM~\ref{SMss:10cl_more} we provide more evidence by focusing on other tasks in between the semantic and the random ones. 

\section{Discussion}\label{s:disc}

CNNs are at the heart of modern machine learning for vision. They have been inspired by our visual system and they have proven their effectiveness in countless tasks. However, there is no \textit{a priori} justification for convolutional architectures, and no reason why other even more effective architectures should not exist. Indeed, several modification of the original convolutional idea have proven effective~\cite{He16,Szegedy16,Yu16}, and recently new architectures such as transformers have challenged the CNN supremacy in vision~\cite{Dosovitskiy20}.

It is therefore interesting to see if the CNN architecture, or something with similar characteristics, could be naturally obtained from data.
A recent work~\cite{Neyshabur20} takes a first important step in this direction, with a custom modification of the LASSO technique~\cite{Tibshirani96}, and working on a small but highly augmented dataset, it shows the emergence in training of local and sparse nonzero weights.
In this work, we have followed a different perspective. With the aim of characterizing the inductive bias induced by pruning, we have limited ourselves to the original inputs, and we have applied iterative magnitude pruning.
This approach has two benefits: it naturally leads to \textit{masks}, i.e.\ it irreversibly changes the architecture of the original network, and it leads to a final model that can be trained \textit{from scratch} to the final accuracy (i.e.\ from the original initialization values, although these are still correlated with the masks).
In combination with the lack of augmentation, this makes features selected from this procedure the natural structure emerging from the dataset and task at hand.

We have presented a thorough analysis of the architecture thus found: the pruning procedure does not affect all nodes equally, but a small number of them retains a disproportionate amount of inputs.
These nodes are particularly important for the performance of the network and they correspond to input masks that are local in space and color channels, with a preference for specific patterns, found throughout the image (although for these images some features are clearly more likely to appear at some specific position than others).
At visual inspection, these features bear a strong resemblance to those found in CNNs~\cite{Zeiler14}, the visual cortex~\cite{Marcelja80,Wolf05} and obtained with $\beta-$LASSO~\cite{Neyshabur20}.
At high pruning, the features become more local and sparse, and they are combined in the next layer with a preference for similar patterns and close location.

All these characteristics align well with the design choices of CNNs: selecting local features and combining them locally is the natural structure emerging from the image classification task---it represents the winning ticket that can be more easily trained.
While other features like weight sharing simply cannot be reproduced in this model, the presence of similar masks at different locations is a sign of translational invariance (or the closest version of it available in these mostly centered images, an experiment with complete translational invariance is presented in SM \ref{SMss:transl}).
In this sense, this procedure could suggest network structures even better aligned with the data from a real dataset, leading to more effective architectures. For instance the sparse sampling, even at this low image resolution, is reminiscent of dilated convolutions~\cite{Yu16}.

As could be expected, only training on a large enough dataset leads to the emergence of these features. Interestingly, the success of the IMP itself seems to be related to their presence, hinting at the different structure of the solution, in line with recent findings about generalization and prunability~\cite{Kuhn21}.
In the same direction, we have shown that locality and defined features disappear when we replace the task with a more vague one, where IMP is also ineffective. Features are thus not a simple consequence of the dataset, but depend on the structure imposed by the task.

There are several directions which are worth exploring in future works. More complex starting architectures and larger or augmented training sets could be explored, to obtain even closer approximations to CNNs or other more effective architectures. The dependence of the architectural properties induced by pruning on the optimization protocol is another interesting aspect.   
For instance, we have found that just training until the beginning of overfitting is sufficient to obtain these features (see SM \ref{SMss:IMPpar}). We have also noticed that precursors are already visible at the early stages of pruning (see video link in SM \ref{SMs:ext}), suggesting the possibility of their early identification and more refined search strategies. 
In addition, it would be interesting to study more advanced pruning algorithms such as~\cite{Evci20,Lin20}, especially iterative or inherently sparse ones that could explore structures corresponding to prohibitively large networks. 
Finally, repeating our analysis for tasks different from visual ones, e.g. for audio~\cite{Oord16} and time series~\cite{Gamboa17}, and study the architectural bias induced by pruning is certainly an interesting direction for future research. This approach could also highlight useful architectures in fields that are still using more standard FCNs such as material modeling~\cite{Behler16}.

In conclusion, our work underlines the effectiveness of pruning as a tool to uncover not only more efficient networks, but also specific architectural properties associated to the structure of the data and relevant for the training task. 
From this point of view, pruning schemes could offer a way to uncover effective new architectures for a wide category of problems.

\begin{ack}
We thank S. d'Ascoli, S. Goldt, M. Refinetti and L. Sagun for useful discussions. FP would especially like to thank E. K\"{u}\c{c}\"{u}kbenli for insightful ideas.

We acknowledge funding from the French government under management of Agence Nationale de la Recherche as part of the ``Investissements d’avenir'' program, reference ANR-19-P3IA-0001 (PRAIRIE 3IA Institute) and from the Simons Foundation collaboration ``Cracking the Glass Problem'' (No. 454935 to G. Biroli). 
\end{ack}

\bibliographystyle{unsrtnat}
\bibliography{FC2CNN}

\begin{thebibliography}{46}
\providecommand{\natexlab}[1]{#1}
\providecommand{\url}[1]{\texttt{#1}}
\expandafter\ifx\csname urlstyle\endcsname\relax
  \providecommand{\doi}[1]{doi: #1}\else
  \providecommand{\doi}{doi: \begingroup \urlstyle{rm}\Url}\fi

\bibitem[Krizhevsky et~al.(2012)Krizhevsky, Sutskever, and Hinton]{Alex12}
Alex Krizhevsky, Ilya Sutskever, and Geoffrey~E Hinton.
\newblock Imagenet classification with deep convolutional neural networks.
\newblock In F.~Pereira, C.~J.~C. Burges, L.~Bottou, and K.~Q. Weinberger,
  editors, \emph{Advances in Neural Information Processing Systems}, volume~25.
  Curran Associates, Inc., 2012.

\bibitem[He et~al.(2016)He, Zhang, Ren, and Sun]{He16}
Kaiming He, Xiangyu Zhang, Shaoqing Ren, and Jian Sun.
\newblock Deep residual learning for image recognition.
\newblock In \emph{Proceedings of the IEEE Conference on Computer Vision and
  Pattern Recognition (CVPR)}, June 2016.

\bibitem[Vaswani et~al.(2017)Vaswani, Shazeer, Parmar, Uszkoreit, Jones, Gomez,
  Kaiser, and Polosukhin]{Vaswani17}
Ashish Vaswani, Noam Shazeer, Niki Parmar, Jakob Uszkoreit, Llion Jones,
  Aidan~N. Gomez, Lukasz Kaiser, and Illia Polosukhin.
\newblock Attention is all you need.
\newblock \emph{arXiv preprint 1706.03762}, 2017.

\bibitem[Brown et~al.(2020)Brown, Mann, Ryder, Subbiah, Kaplan, Dhariwal,
  Neelakantan, Shyam, Sastry, Askell, Agarwal, Herbert-Voss, Krueger, Henighan,
  Child, Ramesh, Ziegler, Wu, Winter, Hesse, Chen, Sigler, Litwin, Gray, Chess,
  Clark, Berner, McCandlish, Radford, Sutskever, and Amodei]{GPT3}
Tom~B. Brown, Benjamin Mann, Nick Ryder, Melanie Subbiah, Jared Kaplan,
  Prafulla Dhariwal, Arvind Neelakantan, Pranav Shyam, Girish Sastry, Amanda
  Askell, Sandhini Agarwal, Ariel Herbert-Voss, Gretchen Krueger, Tom Henighan,
  Rewon Child, Aditya Ramesh, Daniel~M. Ziegler, Jeffrey Wu, Clemens Winter,
  Christopher Hesse, Mark Chen, Eric Sigler, Mateusz Litwin, Scott Gray,
  Benjamin Chess, Jack Clark, Christopher Berner, Sam McCandlish, Alec Radford,
  Ilya Sutskever, and Dario Amodei.
\newblock Language models are few-shot learners.
\newblock \emph{arXiv preprint 2005.14165}, 2020.

\bibitem[Silver et~al.(2017)Silver, Hubert, Schrittwieser, Antonoglou, Lai,
  Guez, Lanctot, Sifre, Kumaran, Graepel, Lillicrap, Simonyan, and
  Hassabis]{Silver17}
David Silver, Thomas Hubert, Julian Schrittwieser, Ioannis Antonoglou, Matthew
  Lai, Arthur Guez, Marc Lanctot, Laurent Sifre, Dharshan Kumaran, Thore
  Graepel, Timothy Lillicrap, Karen Simonyan, and Demis Hassabis.
\newblock Mastering chess and shogi by self-play with a general reinforcement
  learning algorithm.
\newblock \emph{arXiv preprint 1712.01815}, 2017.

\bibitem[Tan and Le(2020)]{Tan19}
Mingxing Tan and Quoc~V. Le.
\newblock Efficientnet: Rethinking model scaling for convolutional neural
  networks.
\newblock \emph{arXiv preprint 1905.11946}, 2020.

\bibitem[Hoefler et~al.(2021)Hoefler, Alistarh, Ben-Nun, Dryden, and
  Peste]{Hoefler21}
Torsten Hoefler, Dan Alistarh, Tal Ben-Nun, Nikoli Dryden, and Alexandra Peste.
\newblock Sparsity in deep learning: Pruning and growth for efficient inference
  and training in neural networks.
\newblock \emph{arXiv preprint 2102.00554}, 2021.

\bibitem[Frankle and Carbin(2019)]{Frankle19}
Jonathan Frankle and Michael Carbin.
\newblock The lottery ticket hypothesis: Finding sparse, trainable neural
  networks.
\newblock In \emph{ICLR}. OpenReview.net, 2019.

\bibitem[{LeCun} et~al.(1989){LeCun}, {Boser}, {Denker}, {Henderson}, {Howard},
  {Hubbard}, and {Jackel}]{LeCun89}
Y.~{LeCun}, B.~{Boser}, J.~S. {Denker}, D.~{Henderson}, R.~E. {Howard},
  W.~{Hubbard}, and L.~D. {Jackel}.
\newblock Backpropagation applied to handwritten zip code recognition.
\newblock \emph{Neural Computation}, 1\penalty0 (4):\penalty0 541--551, 1989.
\newblock \doi{10.1162/neco.1989.1.4.541}.

\bibitem[Simonyan and Zisserman(2015)]{Simonyan15}
Karen Simonyan and Andrew Zisserman.
\newblock Very deep convolutional networks for large-scale image recognition.
\newblock \emph{arXiv preprint 1409.1556}, 2015.

\bibitem[Szegedy et~al.(2015)Szegedy, Liu, Jia, Sermanet, Reed, Anguelov,
  Erhan, Vanhoucke, and Rabinovich]{Szegedy15}
Christian Szegedy, Wei Liu, Yangqing Jia, Pierre Sermanet, Scott Reed, Dragomir
  Anguelov, Dumitru Erhan, Vincent Vanhoucke, and Andrew Rabinovich.
\newblock Going deeper with convolutions.
\newblock In \emph{Proceedings of the IEEE Conference on Computer Vision and
  Pattern Recognition (CVPR)}, June 2015.

\bibitem[d'Ascoli et~al.(2019)d'Ascoli, Sagun, Biroli, and Bruna]{Dascoli19}
St\'{e}phane d'Ascoli, Levent Sagun, Giulio Biroli, and Joan Bruna.
\newblock Finding the needle in the haystack with convolutions: on the benefits
  of architectural bias.
\newblock In H.~Wallach, H.~Larochelle, A.~Beygelzimer, F.~d'Alch\'{e} Buc,
  E.~Fox, and R.~Garnett, editors, \emph{Advances in Neural Information
  Processing Systems}, volume~32. Curran Associates, Inc., 2019.

\bibitem[Hubel and Wiesel(1962)]{Hubel62}
D.~H. Hubel and T.~N. Wiesel.
\newblock Receptive fields, binocular interaction and functional architecture
  in the cat's visual cortex.
\newblock \emph{The Journal of physiology}, 160\penalty0 (1):\penalty0
  106--154, Jan 1962.
\newblock \doi{10.1113/jphysiol.1962.sp006837}.

\bibitem[Fukushima(1980)]{Fukushima80}
Kunihiko Fukushima.
\newblock Neocognitron: A self-organizing neural network model for a mechanism
  of pattern recognition unaffected by shift in position.
\newblock \emph{Biological Cybernetics}, 36\penalty0 (4):\penalty0 193--202,
  April 1980.
\newblock \doi{10.1007/bf00344251}.

\bibitem[Ruderman(1994)]{Ruderman94}
Daniel~L Ruderman.
\newblock The statistics of natural images.
\newblock \emph{Network: Computation in Neural Systems}, 5\penalty0
  (4):\penalty0 517--548, 1994.
\newblock \doi{10.1088/0954-898X\_5\_4\_006}.

\bibitem[Morcos et~al.(2019)Morcos, Yu, Paganini, and Tian]{Morcos19}
Ari~S. Morcos, Haonan Yu, Michela Paganini, and Yuandong Tian.
\newblock One ticket to win them all: generalizing lottery ticket
  initializations across datasets and optimizers.
\newblock \emph{arXiv preprint 1906.02773}, 2019.

\bibitem[Chrabaszcz et~al.(2017)Chrabaszcz, Loshchilov, and Hutter]{ImageNet32}
Patryk Chrabaszcz, Ilya Loshchilov, and Frank Hutter.
\newblock A downsampled variant of imagenet as an alternative to the {CIFAR}
  datasets.
\newblock \emph{arXiv preprint 1707.08819}, 2017.

\bibitem[Zeiler and Fergus(2014)]{Zeiler14}
Matthew~D. Zeiler and Rob Fergus.
\newblock Visualizing and understanding convolutional networks.
\newblock In David Fleet, Tomas Pajdla, Bernt Schiele, and Tinne Tuytelaars,
  editors, \emph{Computer Vision -- ECCV 2014}, pages 818--833, Cham, 2014.
  Springer International Publishing.
\newblock ISBN 978-3-319-10590-1.

\bibitem[Zhou et~al.(2019)Zhou, Lan, Liu, and Yosinski]{Zhou19}
Hattie Zhou, Janice Lan, Rosanne Liu, and Jason Yosinski.
\newblock Deconstructing lottery tickets: Zeros, signs, and the supermask.
\newblock \emph{arXiv preprint 1905.01067}, 2019.

\bibitem[Liu et~al.(2019)Liu, Sun, Zhou, Huang, and Darrell]{Liu19}
Zhuang Liu, Mingjie Sun, Tinghui Zhou, Gao Huang, and Trevor Darrell.
\newblock Rethinking the value of network pruning.
\newblock \emph{arXiv preprint 1810.05270}, 2019.

\bibitem[Gale et~al.(2019)Gale, Elsen, and Hooker]{Gale19}
Trevor Gale, Erich Elsen, and Sara Hooker.
\newblock The state of sparsity in deep neural networks.
\newblock \emph{arXiv preprint 1902.09574}, 2019.

\bibitem[Frankle et~al.(2019)Frankle, Dziugaite, Roy, and Carbin]{Frankle19-2}
Jonathan Frankle, Gintare~Karolina Dziugaite, Daniel~M. Roy, and Michael
  Carbin.
\newblock Linear mode connectivity and the lottery ticket hypothesis.
\newblock \emph{arXiv preprint 1912.05671}, 2019.

\bibitem[Frankle et~al.(2020{\natexlab{a}})Frankle, Dziugaite, Roy, and
  Carbin]{Frankle20}
Jonathan Frankle, Gintare~Karolina Dziugaite, Daniel~M. Roy, and Michael
  Carbin.
\newblock Stabilizing the lottery ticket hypothesis.
\newblock \emph{arXiv preprint 1903.01611}, 2020{\natexlab{a}}.

\bibitem[Frankle et~al.(2020{\natexlab{b}})Frankle, Schwab, and
  Morcos]{Frankle20-2}
Jonathan Frankle, David~J. Schwab, and Ari~S. Morcos.
\newblock The early phase of neural network training.
\newblock \emph{arXiv preprint 2002.10365}, 2020{\natexlab{b}}.

\bibitem[Renda et~al.(2020)Renda, Frankle, and Carbin]{Renda20}
Alex Renda, Jonathan Frankle, and Michael Carbin.
\newblock Comparing rewinding and fine-tuning in neural network pruning.
\newblock \emph{arXiv preprint 2003.02389}, 2020.

\bibitem[Sabatelli et~al.(2020)Sabatelli, Kestemont, and Geurts]{Sabatelli20}
Matthia Sabatelli, Mike Kestemont, and Pierre Geurts.
\newblock On the transferability of winning tickets in non-natural image
  datasets.
\newblock \emph{arXiv preprint 2005.05232}, 2020.

\bibitem[Neyshabur(2020)]{Neyshabur20}
Behnam Neyshabur.
\newblock Towards learning convolutions from scratch.
\newblock \emph{arXiv preprint 2007.13657}, 2020.

\bibitem[Tolstikhin et~al.(2021)Tolstikhin, Houlsby, Kolesnikov, Beyer, Zhai,
  Unterthiner, Yung, Keysers, Uszkoreit, Lucic, et~al.]{tolstikhin2021mlp}
Ilya Tolstikhin, Neil Houlsby, Alexander Kolesnikov, Lucas Beyer, Xiaohua Zhai,
  Thomas Unterthiner, Jessica Yung, Daniel Keysers, Jakob Uszkoreit, Mario
  Lucic, et~al.
\newblock Mlp-mixer: An all-mlp architecture for vision.
\newblock \emph{arXiv preprint arXiv:2105.01601}, 2021.

\bibitem[Russakovsky et~al.(2015)Russakovsky, Deng, Su, Krause, Satheesh, Ma,
  Huang, Karpathy, Khosla, Bernstein, Berg, and Fei-Fei]{ILSVRC15}
Olga Russakovsky, Jia Deng, Hao Su, Jonathan Krause, Sanjeev Satheesh, Sean Ma,
  Zhiheng Huang, Andrej Karpathy, Aditya Khosla, Michael Bernstein,
  Alexander~C. Berg, and Li~Fei-Fei.
\newblock {ImageNet Large Scale Visual Recognition Challenge}.
\newblock \emph{International Journal of Computer Vision (IJCV)}, 115\penalty0
  (3):\penalty0 211--252, 2015.
\newblock \doi{10.1007/s11263-015-0816-y}.

\bibitem[Ioffe and Szegedy(2015)]{Ioffe15}
Sergey Ioffe and Christian Szegedy.
\newblock Batch normalization: Accelerating deep network training by reducing
  internal covariate shift.
\newblock In Francis Bach and David Blei, editors, \emph{Proceedings of the
  32nd International Conference on Machine Learning}, volume~37 of
  \emph{Proceedings of Machine Learning Research}, pages 448--456, Lille,
  France, 07--09 Jul 2015. PMLR.

\bibitem[Glorot and Bengio(2010)]{Glorot10}
Xavier Glorot and Yoshua Bengio.
\newblock Understanding the difficulty of training deep feedforward neural
  networks.
\newblock In Yee~Whye Teh and Mike Titterington, editors, \emph{Proceedings of
  the Thirteenth International Conference on Artificial Intelligence and
  Statistics}, volume~9 of \emph{Proceedings of Machine Learning Research},
  pages 249--256, Chia Laguna Resort, Sardinia, Italy, 13--15 May 2010. PMLR.

\bibitem[You et~al.(2020)You, Li, Xu, Fu, Wang, Chen, Baraniuk, Wang, and
  Lin]{You20}
Haoran You, Chaojian Li, Pengfei Xu, Yonggan Fu, Yue Wang, Xiaohan Chen,
  Richard~G. Baraniuk, Zhangyang Wang, and Yingyan Lin.
\newblock Drawing early-bird tickets: Towards more efficient training of deep
  networks.
\newblock \emph{arXiv preprint 1909.11957}, 2020.

\bibitem[Blalock et~al.(2020)Blalock, Ortiz, Frankle, and
  Guttag]{blalock2020state}
Davis Blalock, Jose Javier~Gonzalez Ortiz, Jonathan Frankle, and John Guttag.
\newblock What is the state of neural network pruning?
\newblock \emph{arXiv preprint arXiv:2003.03033}, 2020.

\bibitem[Mar\^{c}elja(1980)]{Marcelja80}
S.~Mar\^{c}elja.
\newblock Mathematical description of the responses of simple cortical
  cells$\ast$.
\newblock \emph{J. Opt. Soc. Am.}, 70\penalty0 (11):\penalty0 1297--1300, Nov
  1980.
\newblock \doi{10.1364/JOSA.70.001297}.

\bibitem[Szegedy et~al.(2016)Szegedy, Vanhoucke, Ioffe, Shlens, and
  Wojna]{Szegedy16}
Christian Szegedy, Vincent Vanhoucke, Sergey Ioffe, Jon Shlens, and Zbigniew
  Wojna.
\newblock Rethinking the inception architecture for computer vision.
\newblock In \emph{Proceedings of the IEEE Conference on Computer Vision and
  Pattern Recognition (CVPR)}, June 2016.

\bibitem[Yu and Koltun(2016)]{Yu16}
Fisher Yu and Vladlen Koltun.
\newblock Multi-scale context aggregation by dilated convolutions, 2016.

\bibitem[Dosovitskiy et~al.(2020)Dosovitskiy, Beyer, Kolesnikov, Weissenborn,
  Zhai, Unterthiner, Dehghani, Minderer, Heigold, Gelly, Uszkoreit, and
  Houlsby]{Dosovitskiy20}
Alexey Dosovitskiy, Lucas Beyer, Alexander Kolesnikov, Dirk Weissenborn,
  Xiaohua Zhai, Thomas Unterthiner, Mostafa Dehghani, Matthias Minderer, Georg
  Heigold, Sylvain Gelly, Jakob Uszkoreit, and Neil Houlsby.
\newblock An image is worth 16x16 words: Transformers for image recognition at
  scale.
\newblock \emph{arXiv preprint 2010.11929}, 2020.

\bibitem[Tibshirani(1996)]{Tibshirani96}
Robert Tibshirani.
\newblock Regression shrinkage and selection via the lasso.
\newblock \emph{Journal of the Royal Statistical Society: Series B
  (Methodological)}, 58\penalty0 (1):\penalty0 267--288, 1996.
\newblock \doi{https://doi.org/10.1111/j.2517-6161.1996.tb02080.x}.

\bibitem[Wolf et~al.(2005)Wolf, Serre, and Poggio]{Wolf05}
L.~Wolf, T.~Serre, and T.~Poggio.
\newblock Object recognition with features inspired by visual cortex.
\newblock In \emph{Proceedings. 2005 IEEE Computer Society Conference on
  Computer Vision and Pattern Recognition}, volume~3, pages 994--1000, Los
  Alamitos, CA, USA, jun 2005. IEEE Computer Society.
\newblock \doi{10.1109/CVPR.2005.254}.

\bibitem[Kuhn et~al.(2021)Kuhn, Lyle, Gomez, Rothfuss, and Gal]{Kuhn21}
Lorenz Kuhn, Clare Lyle, Aidan~N. Gomez, Jonas Rothfuss, and Yarin Gal.
\newblock Robustness to pruning predicts generalization in deep neural
  networks.
\newblock \emph{arXiv preprint 2103.06002}, 2021.

\bibitem[Evci et~al.(2020)Evci, Gale, Menick, Castro, and Elsen]{Evci20}
Utku Evci, Trevor Gale, Jacob Menick, Pablo~Samuel Castro, and Erich Elsen.
\newblock Rigging the lottery: Making all tickets winners.
\newblock In \emph{Proceedings of the 37th International Conference on Machine
  Learning}, pages 471--481, 2020.

\bibitem[Lin et~al.(2020)Lin, Stich, Barba, Dmitriev, and Jaggi]{Lin20}
Tao Lin, Sebastian~U. Stich, Luis Barba, Daniil Dmitriev, and Martin Jaggi.
\newblock Dynamic model pruning with feedback.
\newblock \emph{arXiv preprint 2006.07253}, 2020.

\bibitem[van~den Oord et~al.(2016)van~den Oord, Dieleman, Zen, Simonyan,
  Vinyals, Graves, Kalchbrenner, Senior, and Kavukcuoglu]{Oord16}
Aaron van~den Oord, Sander Dieleman, Heiga Zen, Karen Simonyan, Oriol Vinyals,
  Alex Graves, Nal Kalchbrenner, Andrew Senior, and Koray Kavukcuoglu.
\newblock Wavenet: A generative model for raw audio.
\newblock \emph{arXiv preprint 1609.03499}, 2016.

\bibitem[Gamboa(2017)]{Gamboa17}
John Cristian~Borges Gamboa.
\newblock Deep learning for time-series analysis.
\newblock \emph{arXiv preprint 1701.01887}, 2017.

\bibitem[Behler(2016)]{Behler16}
J\"{o}rg Behler.
\newblock Perspective: Machine learning potentials for atomistic simulations.
\newblock \emph{The Journal of Chemical Physics}, 145\penalty0 (17):\penalty0
  170901, 2016.
\newblock \doi{10.1063/1.4966192}.

\bibitem[Kingma and Ba(2014)]{Kingma14}
Diederik~P. Kingma and Jimmy Ba.
\newblock Adam: A method for stochastic optimization.
\newblock \emph{arXiv preprint 1412.6980}, 2014.

\end{thebibliography}

\appendix
\clearpage
\section{Convergence tests}\label{SMs:conv}
In this section we report the result of modifying different hyperparameters of the training or the pruning procedure, to show the robustness of the results presented in the main text.

\subsection{Minimization}\label{SMss:min}
\begin{figure}[!htb]
  \centering
  \includegraphics[width=\columnwidth]{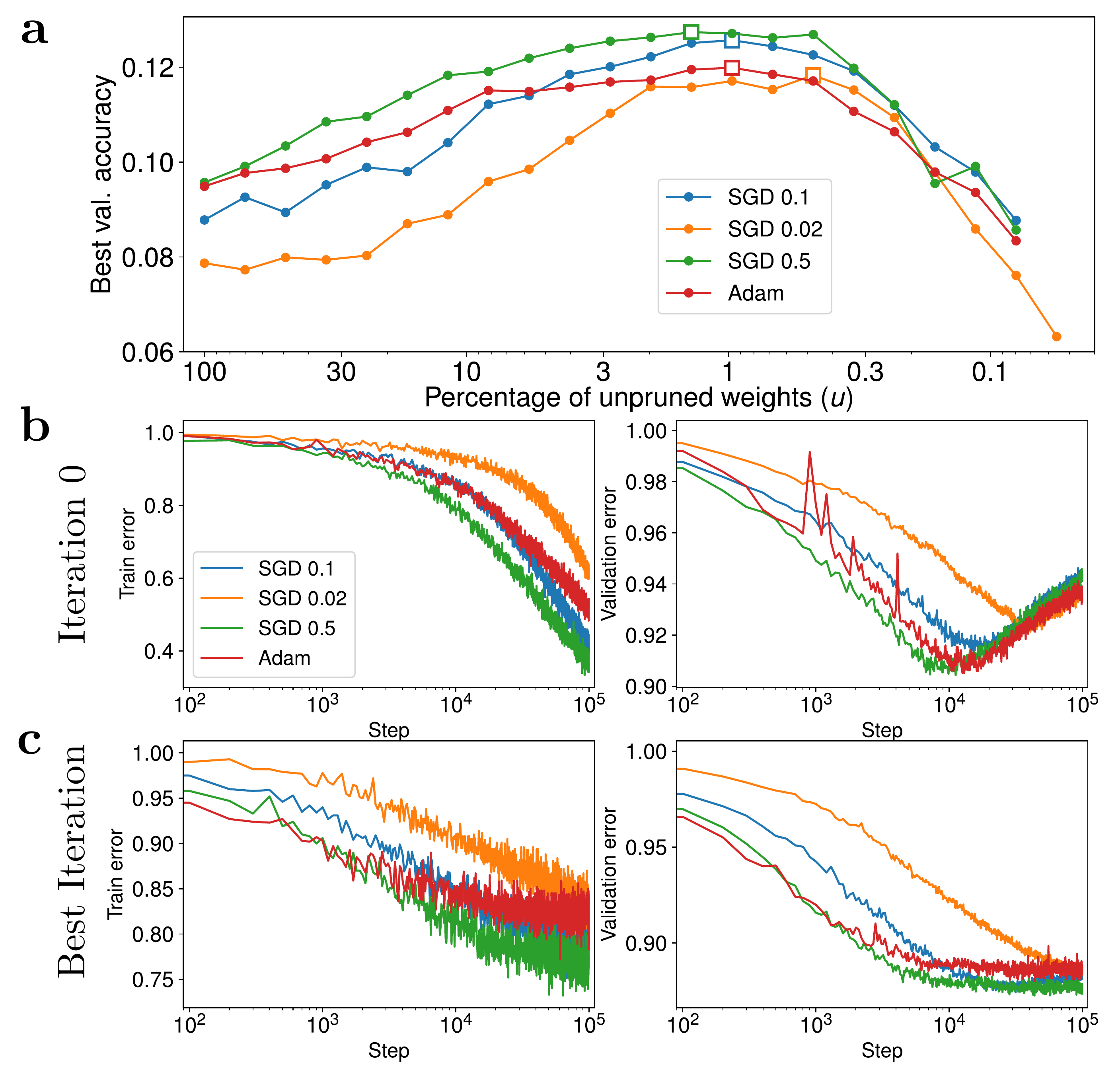}
  \caption{
    \textbf{a:} Best validation accuracy as a function of $u$ for different minimizers: stochastic gradient descent with varying learning rate (SGD in the legend, followed by the learning rate) and Adam with standard parameters and learning rate $0.01$.
    \textbf{b:} Training error (left) and validation error (right) for the initial dense iteration for all minimizers.
    \textbf{c:} Training error (left) and validation error (right) for the best iteration of each minimizer (square in panel \textbf{a}). While training is much slower, all models are able to reach stationary validation within $10^5$ steps, apart from learning rate 0.02 that barely reaches it.
  }\label{fig:mins}
\end{figure}

Fig.~\ref{fig:mins} shows the best validation and training curves when modifying the optimization procedure either by changing the learning rate or by employing a more advanced minimizer (Adam~\cite{Kingma14}).
While there are small differences in the absolute accuracy, the main behavior is robust with respect to the minimization procedure.
Although the value $0.5$ for the learning rate leads to slightly better accuracy, we have employed a more conservative $0.1$ throughout all experiments in the main text, as the final accuracy is not the main scope of this work.
\begin{figure}[!htb]
  \centering
  \includegraphics[width=.8\columnwidth]{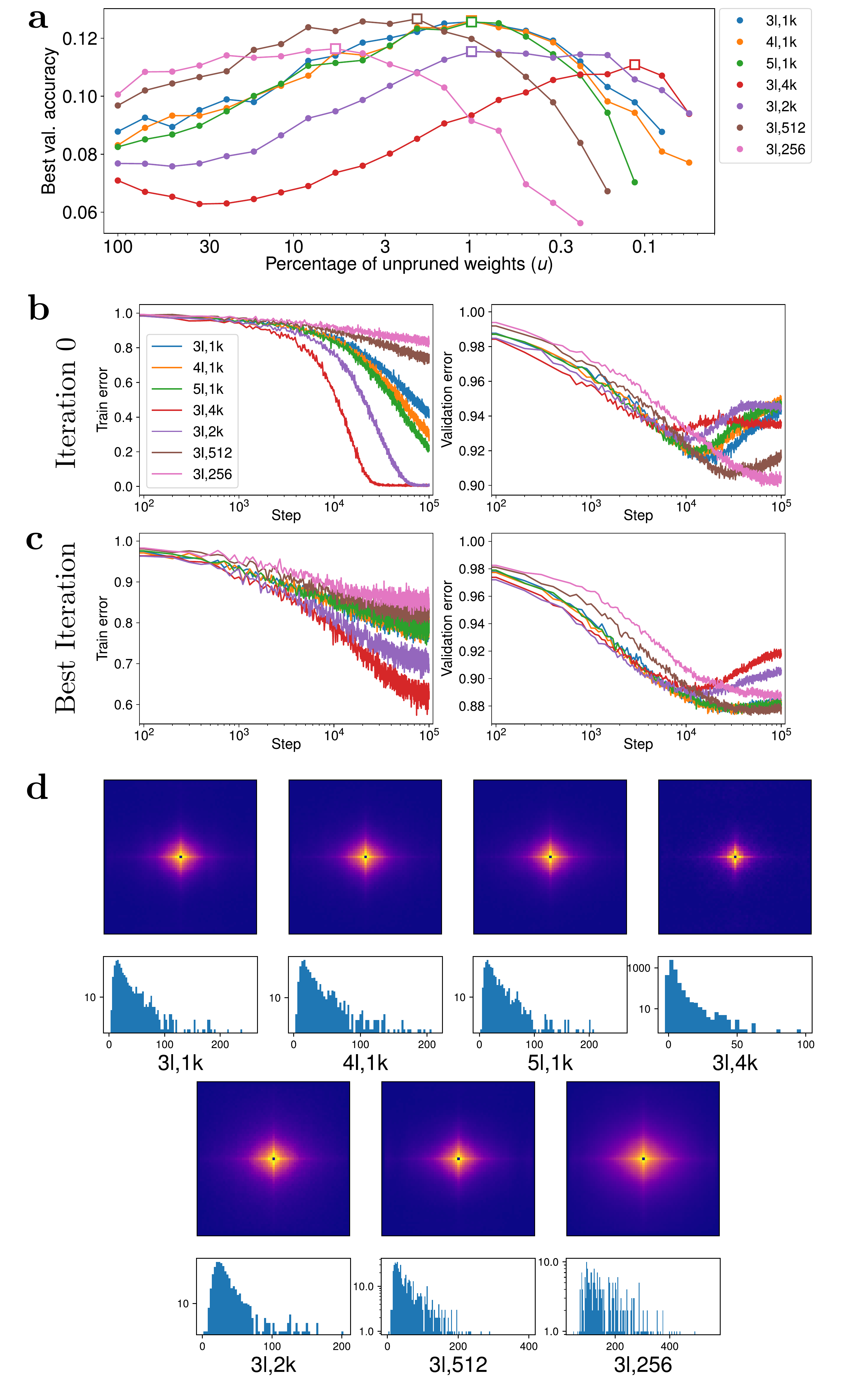}
  \caption{
    \textbf{a:} Best validation accuracy as a function of $u$ for different architectures (see legend and text for notation).
    \textbf{b:} Training error (left) and validation error (right) for the initial dense iteration for all architectures.
    \textbf{c:} Training error (left) and validation error (right) for the best iteration of each architecture (square in panel \textbf{a}).
    \textbf{d:} $S^{\rm sc}$ (see Fig.~\ref{fig:2} for details) and histograms of $C^{{\rm in},1}$ for the best iteration of each architecture (square in panel \textbf{a}).
  }\label{fig:archs}
\end{figure}
\clearpage
\subsection{Other architectures}\label{SMss:arch}
In this section, we modify the network architecture from the original 3 hidden layers of size 1024 (3l, 1k): we deepen the network with 4 (4l, 1k) or 5 (5l, 1k) layers of the same size, or we change the width of each layer (uniformly) to wider 2048 (3l, 2k) and 4096 (3l, 4k), or smaller 512 (3l, 512) and 256 (3l, 256).
All training parameters are kept the same in all these experiments, although better results could be achieved with better minimization and at least the smallest (3l, 256) network could benefit from longer training.

Fig.~\ref{fig:archs}\textbf{a-c} shows the best validation and training curves for each architecture. We see that, at least with these parameters, deepening the network does not have a big effect. For the width: a moderately narrower network is as effective, but larger ones are slightly worse. This might be due to the training procedure, but the difference in performance is already present for the dense networks, and the first pruning steps are not as effective in this case, so these networks might be too overparametrized, leading to a qualitatively different solution.
Panel \textbf{d} shows the locality and connection histogram for the first layer of each network at the best step, indicating that the features discussed in the main text (locality and excess connections nodes) are present in all the cases.

\subsection{IMP parameters}\label{SMss:IMPpar}
We now consider the parameters related to the IMP procedure.
In particular, we act separately on 3 fronts: decreasing the pruning ratio per step ($p=0.2$), decreasing the rewind time for the weights (500, 100 or 0) and varying the total training time per iteration, i.e.\ the time at which we consider the weights for updating the pruning mask ($5\cdot10^5$, $10^4$ and $5\cdot10^3$).

\begin{figure}[!htb]
  \centering
  \includegraphics[width=.8\columnwidth]{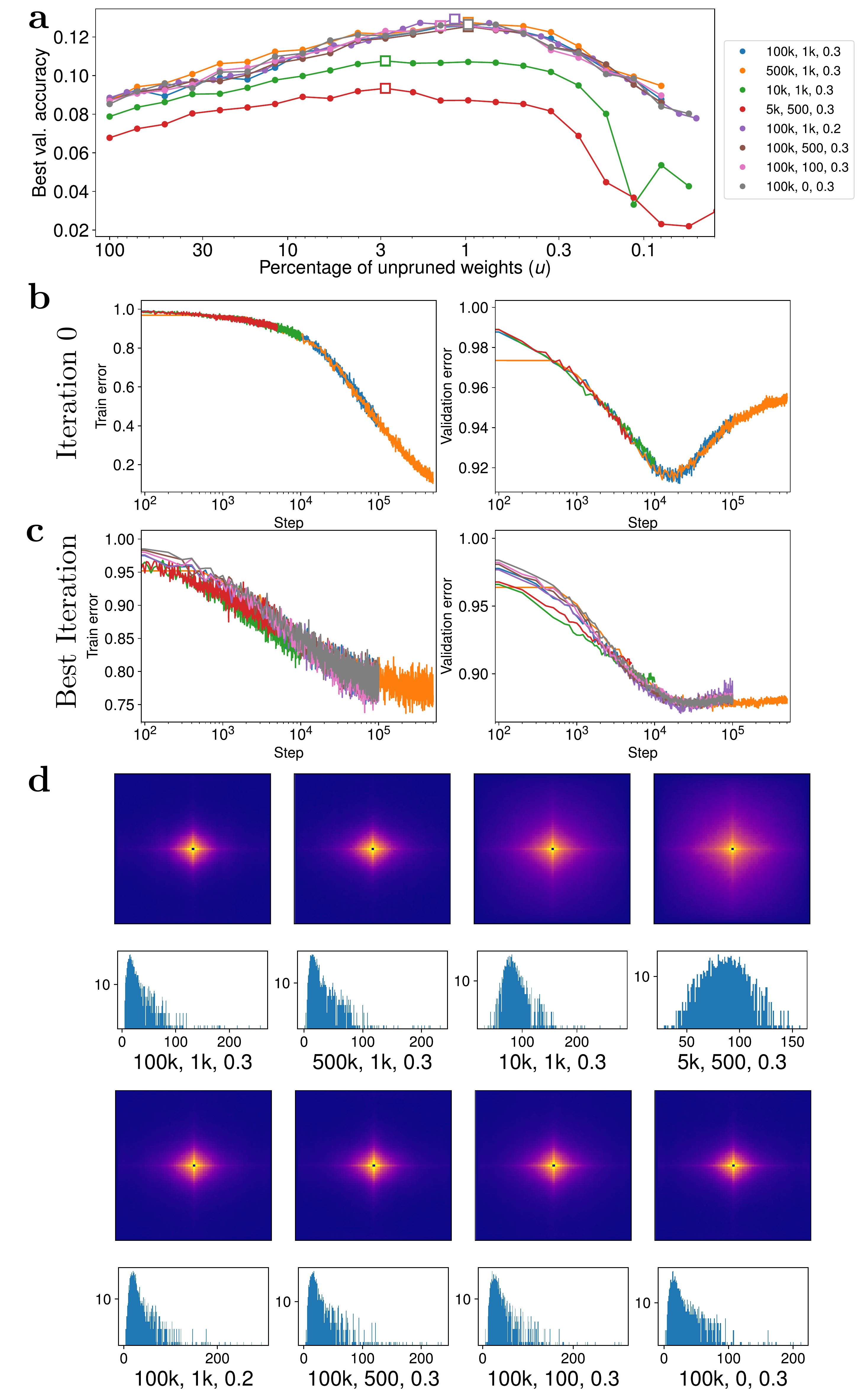}
  \caption{
    \textbf{a:} Best validation accuracy as a function of $u$ for different IMP parameters (train time, rewind time, $p$ in legend).
    \textbf{b:} Training error (left) and validation error (right) for the initial dense iteration for experiments of different total steps.
    \textbf{c:} Training error (left) and validation error (right) for the best iteration of each experiments (square in panel \textbf{a}).
    \textbf{d:} $S^{\rm sc}$ (see Fig.~\ref{fig:2} for details) and histograms of $C^{{\rm in},1}$ for the best iteration of each experiments (square in panel \textbf{a}).
  }\label{fig:IMP}
\end{figure}

Fig.~\ref{fig:IMP}\textbf{a} shows the best validation for each of these experiments (train time, rewind time, $p$ in legend). We can see that most curves lead to the same accuracy, apart from training times shorter than the validation minimum $\sim 2\cdot10^4$, which is unsurprising.
This justifies our practice of stopping training before complete training set overfitting, since the longer run does not offer any benefit. The pruning ratio and rewind time are also clearly converged, we have decided to keep a nonzero rewind time to be more conservative for the other experiments, which could be more brittle.

The other panels of the same figure show dense and best iteration training curves and locality and connectivity plots.
The only cases showing a deviation from the main text results are once again the very short trains. Interestingly, some of the feature seem to survive even for these cases where the network is not even allowed to reach the best validation in each run, suggesting that some of the relevant features are already present in the early stages of training.

\clearpage
\section{More connectivity plots}\label{SMs:masks}
\begin{figure}[!htb]
  \centering
  \includegraphics[width=\columnwidth]{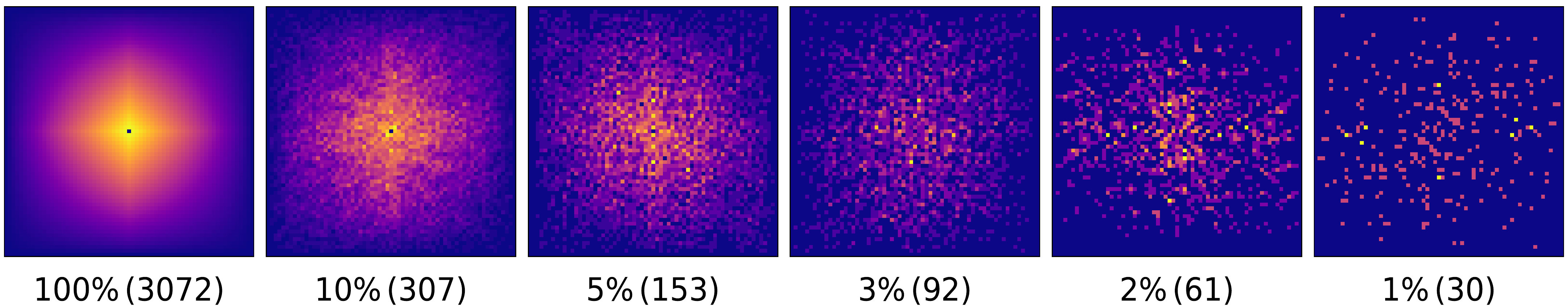}
  \caption{
    $S^{\rm sc}$ for single nodes randomly pruned at a given $u$ (number of surviving connection in parenthesis, see Fig.~\ref{fig:2} for visualization details).
  }\label{fig:ref1}
\end{figure}

In this section we present more specific plots of $S^{\rm sc}(\db)$ and masks for more experiments.
As a reference, Fig.~\ref{fig:ref1} presents the $S^{\rm sc}$ plot for single nodes pruned at random at different sparsity levels.
Even for a fully connected node, the finite size of the image makes local connections more likely than farther connections, but with a decay clearly slower than what shown for localized nodes.
For higher sparsity, we see more and more delocalized patterns appearing (of course this is true for single nodes, while the sum of many random sparse nodes would eventually approximate the fully connected one).

\begin{figure}[!htb]
  \centering
  \includegraphics[width=\columnwidth]{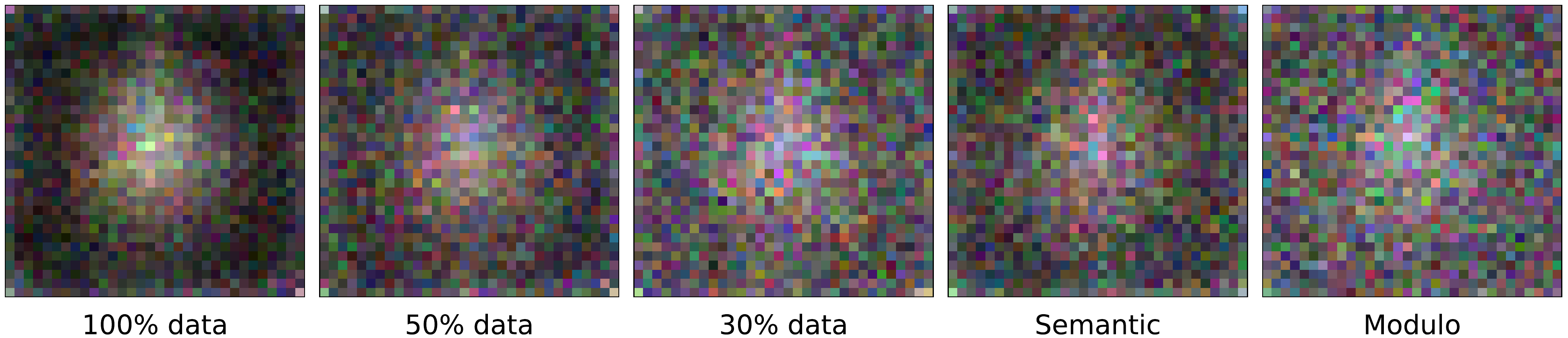}
  \caption{
    $C^{{\rm out},0}$ for different experiments proposed in the main text, at $u\sim 1\%$.
  }\label{fig:ref2}
\end{figure}

Another interesting quantity, shown in Fig.~\ref{fig:ref2} for different networks, is the plot of $C^{{\rm out},0}$, i.e.\ the amount of connections surviving to each pixel. 
We can see that for well performing networks the connectivity is higher to the center of the image, with some extra connections to the edges and corners, while less effective networks tend to have more uniform connectivity.

\subsection{Locality per number of connections}\label{SMss:locconn}
\begin{figure}[!htb]
  \centering
  \includegraphics[width=\columnwidth]{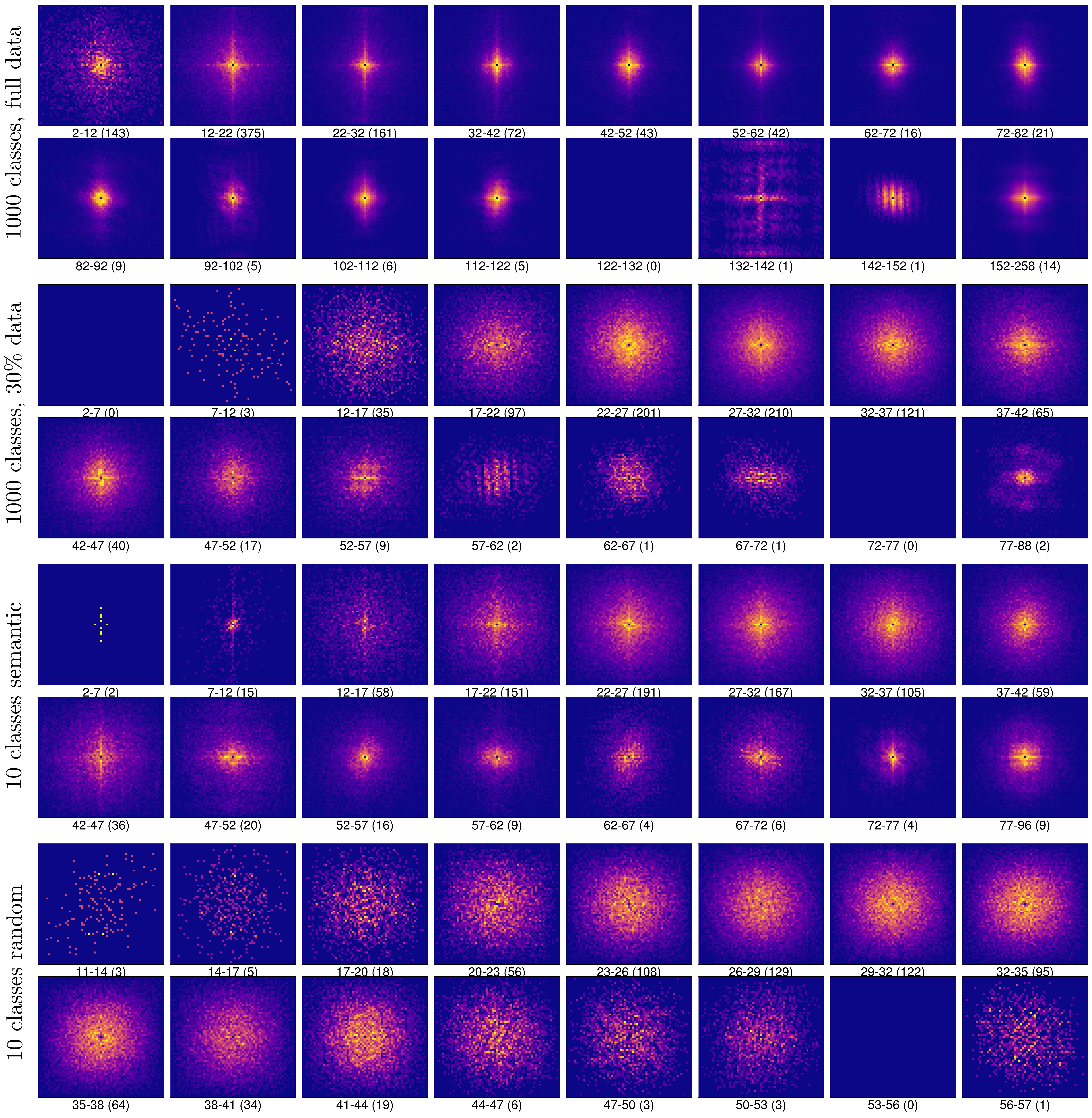}
  \caption{
    $S^{\rm sc}$ of connections to nodes with different values of $C^{{\rm in},1}$.
    From top to bottom: standard network, training on 30\% data, training on 10 semantic classes and training on 10 random classes. All networks are taken at $u\sim1\%$
    Each plot represents the sum of all nodes in an interval of $C^{{\rm in},1}$ shown below the image, with the number of nodes falling in the bin in parenthesis. 
    Intervals are of 10, 5, 5, and 3 bins respectively, to cover the range of connectivity of each network. The last bin covers all remaining nodes.
  }\label{fig:loc}
\end{figure}

Fig.~\ref{fig:loc} presents plots of $S^{\rm sc}$ when we restrict the computation to first hidden layer nodes having values of $C^{{\rm in},1}_j$ in a given range.
This can give us further insight on whether the locality of a network is due to many close pixels in the more pruned nodes, or few localized clusters in the more connected ones.
We present graphs for the standard network, one trained on 30\% of the data and the two versions (semantic and random) of the experiments on 10 classes. All networks are at the same level of pruning $u\sim1\%$.

For the initial network, we see most bins showing some form of locality, more apparent in the most connected bins (some of the later bins contain only a few nodes, strongly affecting the shape of the distance plot).
But for the smaller dataset, we see that only some of the bins with more connections seem to show proper locality, with the less connected ones no better than random. A similar behavior is seen in the network trained on 10 semantic classes. In both these cases features are still partially visible, but we can see that the structure is not preserved in all nodes.
Finally, the network trained on 10 random classes does not show any locality at any scale.

\clearpage
\subsection{Highly connected masks}\label{SMss:bestm}
\begin{figure}[!htb]
  \centering
  \includegraphics[width=\columnwidth]{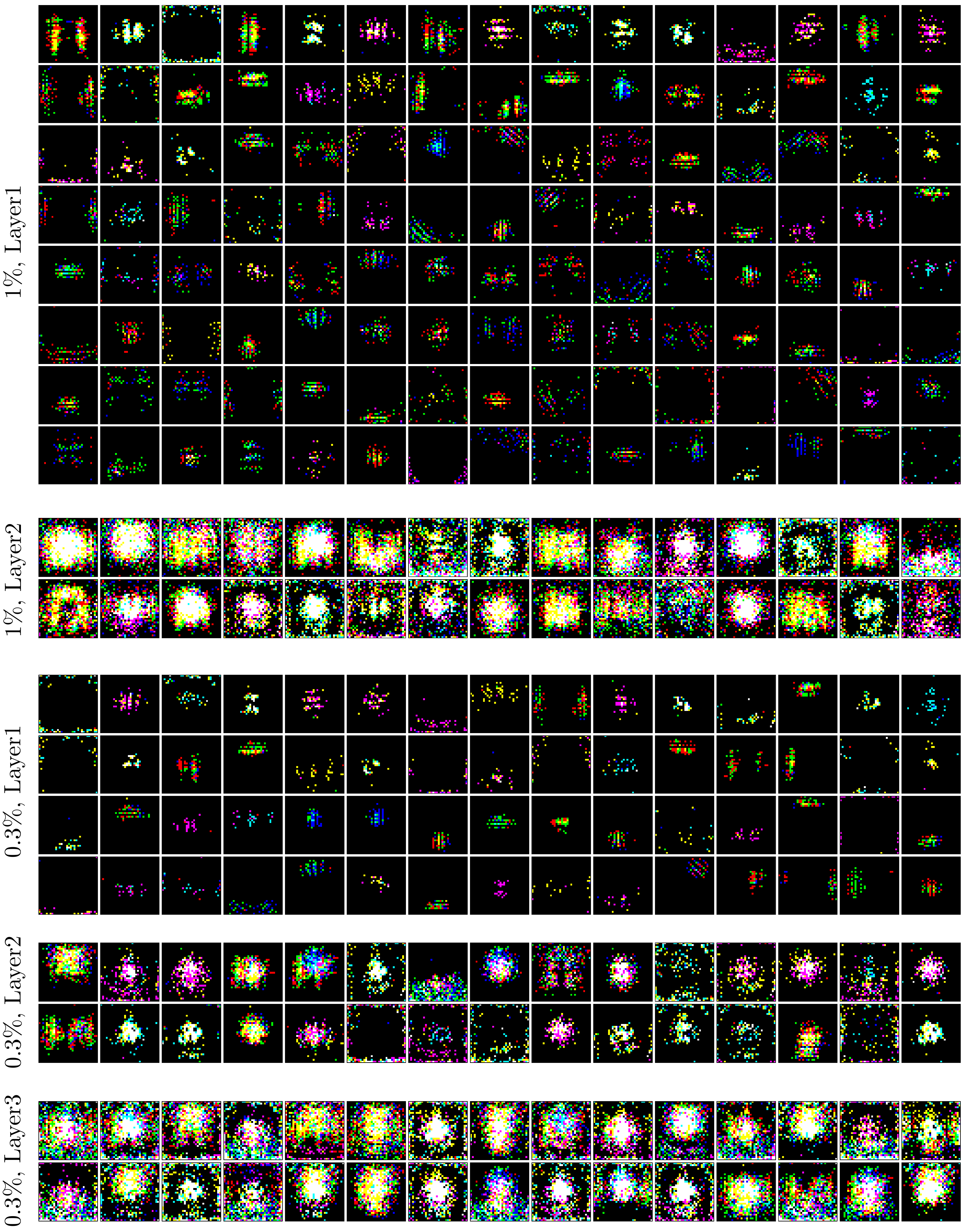}
  \caption{
    Masks of the most connected nodes for different layers and sparsity of the original network (same representation as Fig.~\ref{fig:3}).
  }\label{fig:mask1}
\end{figure}

We show here a larger sample of masks of the most connected nodes.
Fig.~\ref{fig:mask1} covers the original network of Fig.~\ref{fig:1}: for the best sparsity $u\sim1\%$ we show the top 120 most connected layer 1 nodes, showing the variety of features sifted out, and the top 30 effective masks $\mu_{ij}$ of layer 2, showing how they cover most of the image.
For the higher sparsity $u\sim0.3\%$ we show the 60 most connected of the first layer, highlighting how features are preserved, although more local and with higher sparsity. We also show the effective masks of layer 2, partially local, and of layer 3, covering most of the image.

\begin{figure}[!htb]
  \centering
  \includegraphics[width=\columnwidth]{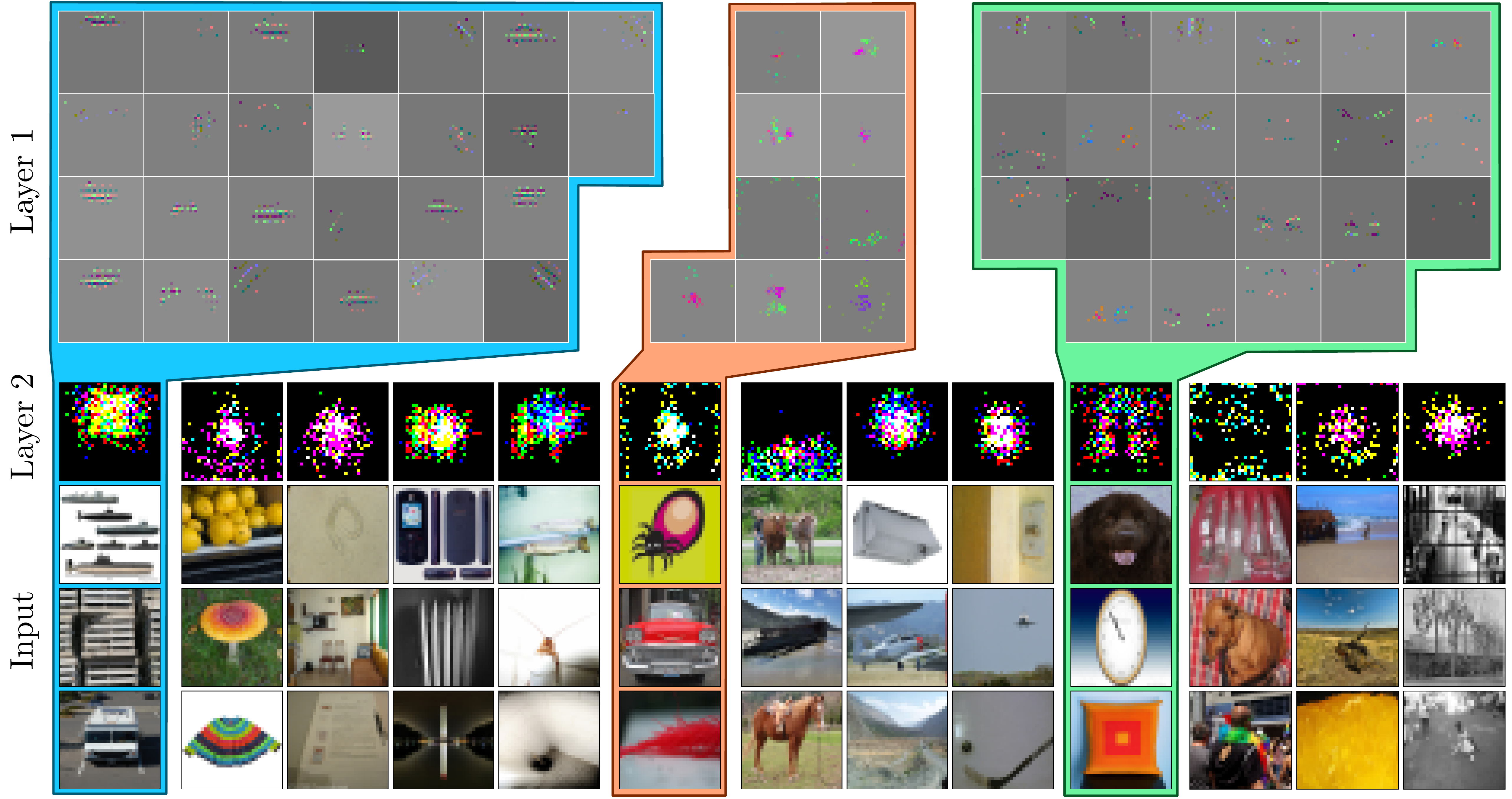}
  \caption{
    Masks of the second layer from input $\mu_{ij}=\theta(\sum_k m^1_{ik}m^2_{kj}-1/2)$ with most nonzero element (center, marked ``Layer 2''), images with the largest activation on them (below, marked ``Input'') and for 3 of them masks in the first layer connected to them (top, marked ``Layer 1''), showing the masked weights $w^1_{ij}m^1_{ij}$ as detailed in Fig.~\ref{fig:3}.
  }\label{fig:masksex4}
\end{figure}

For the same network, at $u\sim0.3\%$, Fig.~\ref{fig:masksex4} shows a more in depth analysis of the masks $\mu_{ij}$ for the most connected nodes of the second hidden layer.
The masks are shown in the center of the image and below them are the 3 images of the validation set that activate the nodes the most.
For 3 sample nodes, the top of the image shows explicitly all the first hidden layer nodes connected to them: they have rather uniform features and a specific local spatial distribution. 
Although not as clean as in a real CNN, we can see how the first layer highlights local features, that are aggregated in a larger but still localized fashion in the second layer, often aggregating similar features at different position, like convolutions preferring a specific channel.

\begin{figure}[!htb]
  \centering
  \includegraphics[width=\columnwidth]{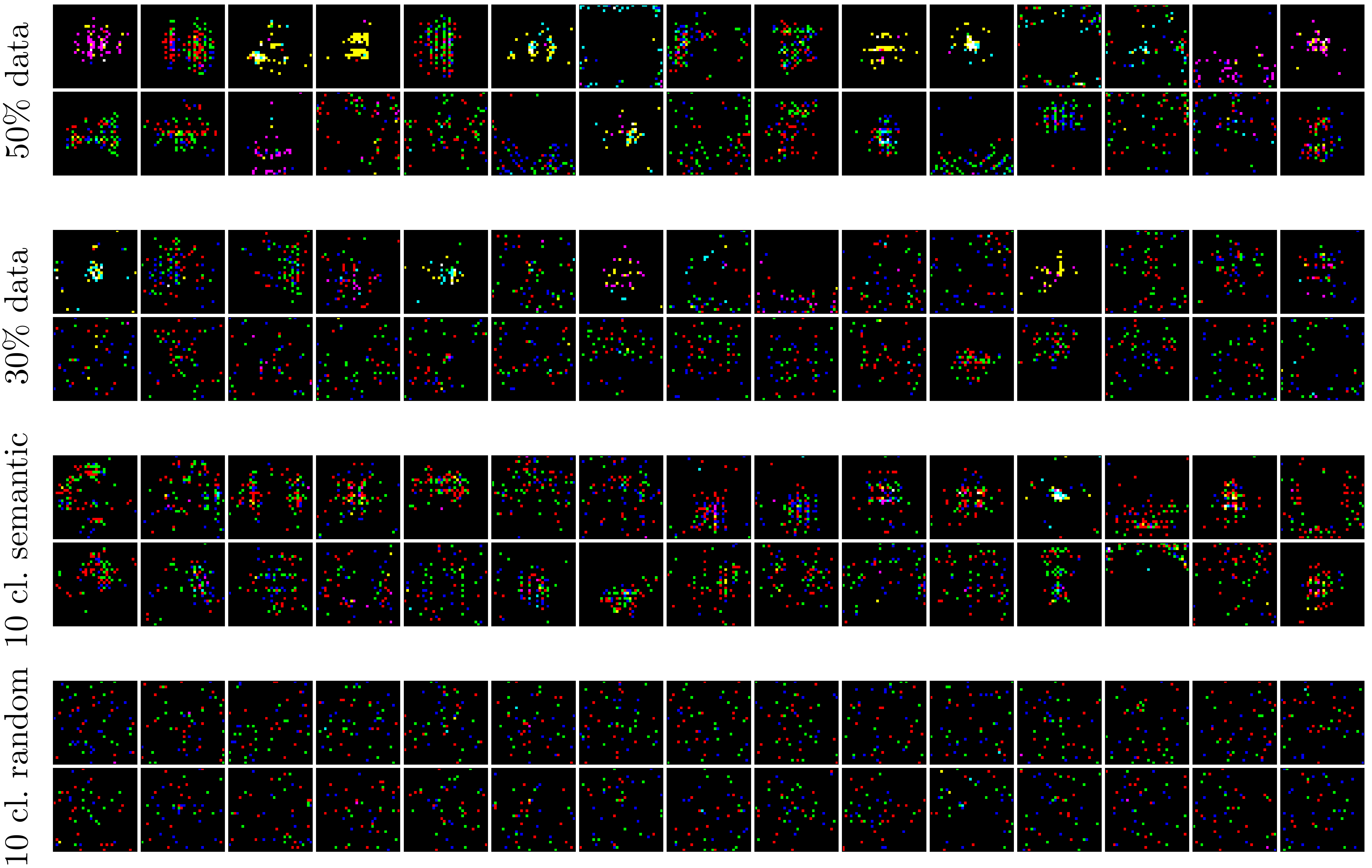}
  \caption{
    Masks of the most connected nodes for different experiments: original task on 50\% and 30\% of the data, and task modified to 10 super-classes either semantically or randomly grouped.
  }\label{fig:mask2}
\end{figure}

Fig.~\ref{fig:mask2} shows the 30 most connected masks in the first layer for other networks: trained on 50\% of the data, 30\% of the data, 10 semantic classes or 10 random classes.
We can see more and more featureless masks among the most connected as we shrink the dataset. For the semantic classes, we still find local structures, although less clear than the original task, while the random classes show no structure at all.

\begin{table}[!b]
    \centering
    \begin{tabular}{c|p{8cm}|c|c}
    \textbf{\#}& \textbf{Class composition (rough)}&\textbf{\#Classes}&\textbf{\#Elements } \rule[-0.9ex]{0pt}{0pt}\\
    \hline
    0& Dog& 118& 147873   \rule{0pt}{2.6ex}\\
    1& Mammal\textbackslash Dog& 100& 129728\\
    2& Bird, Reptile& 95& 122895\\
    3& Fish, Amphibian, Invertebrate& 85& 110034\\
    4& Device\textbackslash Musical instrument& 104& 131965\\
    5& Covering, Musical instrument& 111& 142423\\
    6& Container\textbackslash Vehicle, Appliance, Equipment& 99& 126603\\
    7& Transport, Furnishing& 95& 122661\\
    8& Plant, Fungus, Food, Vegetable, Fruit, Implement& 96& 122701\\
    9& Construction, Miscellaneous& 97& 124282  \rule[-0.9ex]{0pt}{0pt}\\
    \hline
    &Total&1000&1281167   \rule{0pt}{2.6ex}
    \end{tabular}
    \caption{
      Proposed 10 aggregated ImageNet classes: for each class we give a rough definition of the class in terms of macro-categories, the number of original ImageNet classes and single images in those classes that fall into this class (the symbol A\textbackslash B indicates all classes that fall in the macro class A but not in B).
    }
    \label{tab:10cl}
\end{table}

\section{Class clustering}\label{SMs:clust}
We give here more details about the construction of 10 \textit{semantic} super-classes.
Our goal was to design 10 classes that would: include all the original 1000, be similarly sized, and be easily identified in terms of larger WordNet categories.

The details of the final classes constructed are summarized in Table~\ref{tab:10cl}. 
The animals are divided into 4 categories (the large amount of dogs naturally falls into a slightly larger category).
Objects make up most of the remaining 6 classes roughly based on sub-categories of ``artifacts'': device, covering (mostly clothes), container (excluding vehicles), transport and furnishing, implement and construction.
Class 8 also contains all plant-derived categories and food.
Class 9 also includes all categories not included in one of the previous macro categories, listed as miscellaneous (including e.g.\ some natural formations).
A few classes (21) are included in more than 1 macro class by this division, and they were arbitrarily assigned to a single one.
The specific attribution of each of the original classes is available in the code repository specified in \ref{SMs:ext}.

\section{Modified tasks}\label{SMs:otask}
We report here some experiments on modified datasets, that reinforce some of the statements of the main text

\subsection{More super-classes}\label{SMss:10cl_more}
We consider more ways of creating 10 macro classes from the original one. To create a task in between the ``semantic'' and ``random'' ones of Sec.~\ref{s:roletask}, we consider 2 options: one we call ``5 semantic/5 random'', where the first 5 classes are the ones described in Sec.~\ref{SMss:10cl_more}, while the other ones are random groupings of the remaining categories.
The other we call ``partial semantic'' where each of our super-classes is composed of half (around 50) of the categories of the ``semantic'' classes and the other half picked at random between the remaining categories.

\begin{figure}[!htb]
  \centering
  \includegraphics[width=\columnwidth]{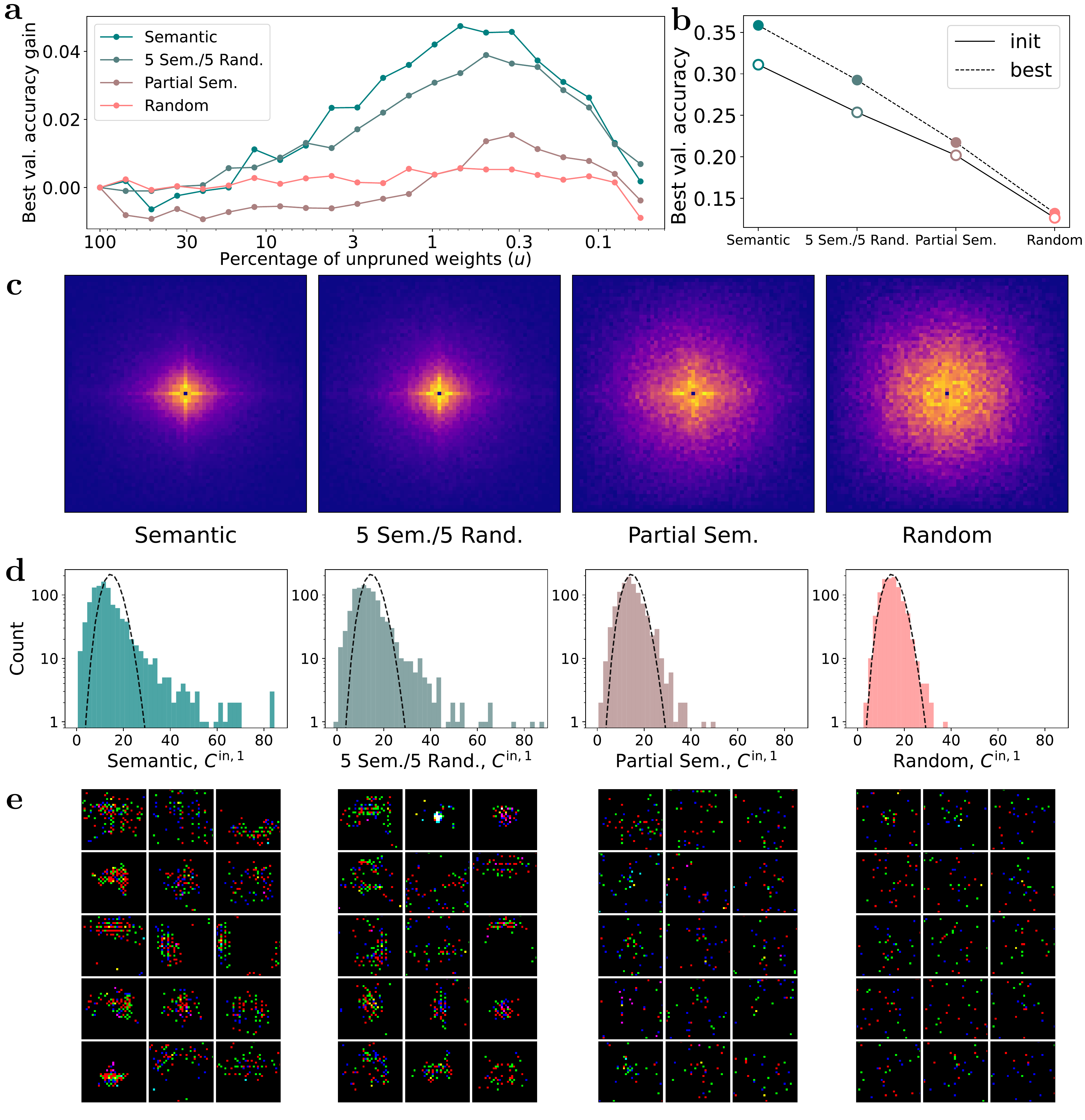}
  \caption{
    \textbf{a:} Gain in best validation accuracy (difference from the value corresponding to the unpruned network) for tasks from semantic to random.
    \textbf{b:} Best validation accuracy for the unpruned network (empty dots) and best IMP iteration (full dots) for the different tasks.
    \textbf{c:} $S^{\rm sc}(\db)$ for the four tasks at the iteration $u\simeq0.5\%$. Normalization and color map as shown in Fig.~\ref{fig:2}.
    \textbf{c:} Histogram of $C^{{\rm in},1}$ for the four tasks at the iteration $u\simeq0.5\%$. The dashed line is the theoretical binomial distribution for random pruning.
    \textbf{e:} Most connected first layer masks for the four tasks.
  }\label{fig:10cl_more}
\end{figure}

To gauge the complexity of the task, Fig.~\ref{fig:10cl_more}\textbf{b} reports the unpruned and best accuracy for each of these experiments, confirming that they are in between the two extremes, the one with 5 ``fully meaningful'' classes being simpler than the other.
The full best validation curves (minus the unpruned values) reported in panel \textbf{a} and the locality and connectivity plots in panels \textbf{c} and \textbf{d} support the idea of a gradual progression between a network characterized by local structures and one where the task is performed in a different way.
Lastly, the most connected first layer masks in panel \textbf{e} highlight the progressive disappearance of features.

\subsection{Rotated images}\label{SMss:rot}
\begin{figure}[!htb]
  \centering
  \includegraphics[width=\columnwidth]{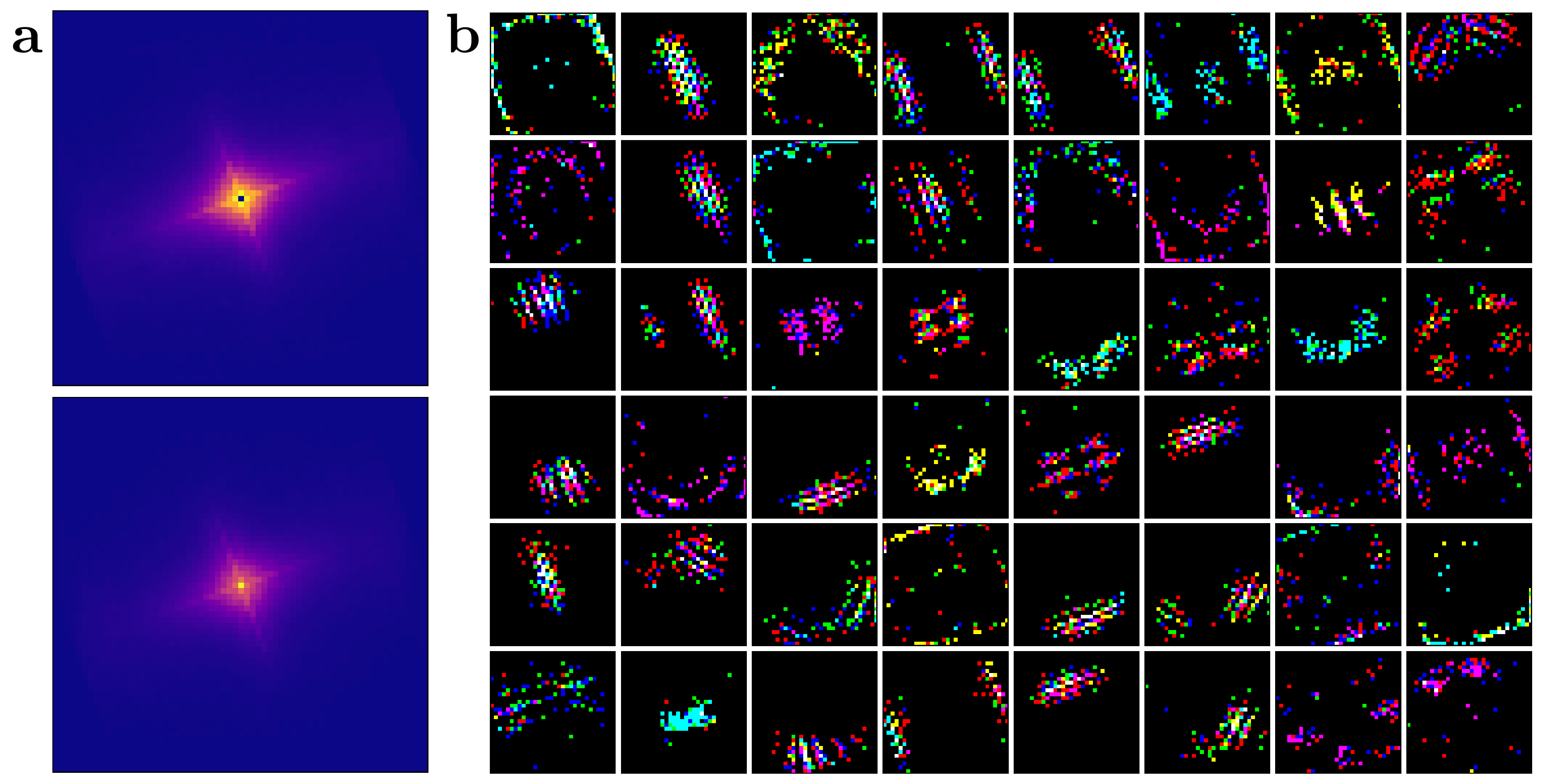}
  \caption{
    \textbf{a:} $S^{\rm sc}$ and $S^{\rm dc}$ for the best iteration of the experiment on rotated images.
    \textbf{b:} masks of the 48 most connected layer 1 nodes for the same iteration.
  }\label{fig:rot}
\end{figure}

We rotate all images 20 degrees counterclockwise, and crop to obtain new $32\times32$ images (we mask out the pixels not corresponding to any original pixels).
This is done to verify that the vertical and horizontal directions have nothing to do with the discretization of the images.
The overall training is similar to the original one, with slightly lower accuracy.
Fig.~\ref{fig:rot} shows $S(\db)$ and masks at the best iteration.
As expected, the preferred directions are rotated and masks tend to follow the new horizontal and vertical alignment.

\subsection{ImageNet64}\label{SMss:IN64}
\begin{figure}[!htb]
  \centering
  \includegraphics[width=\columnwidth]{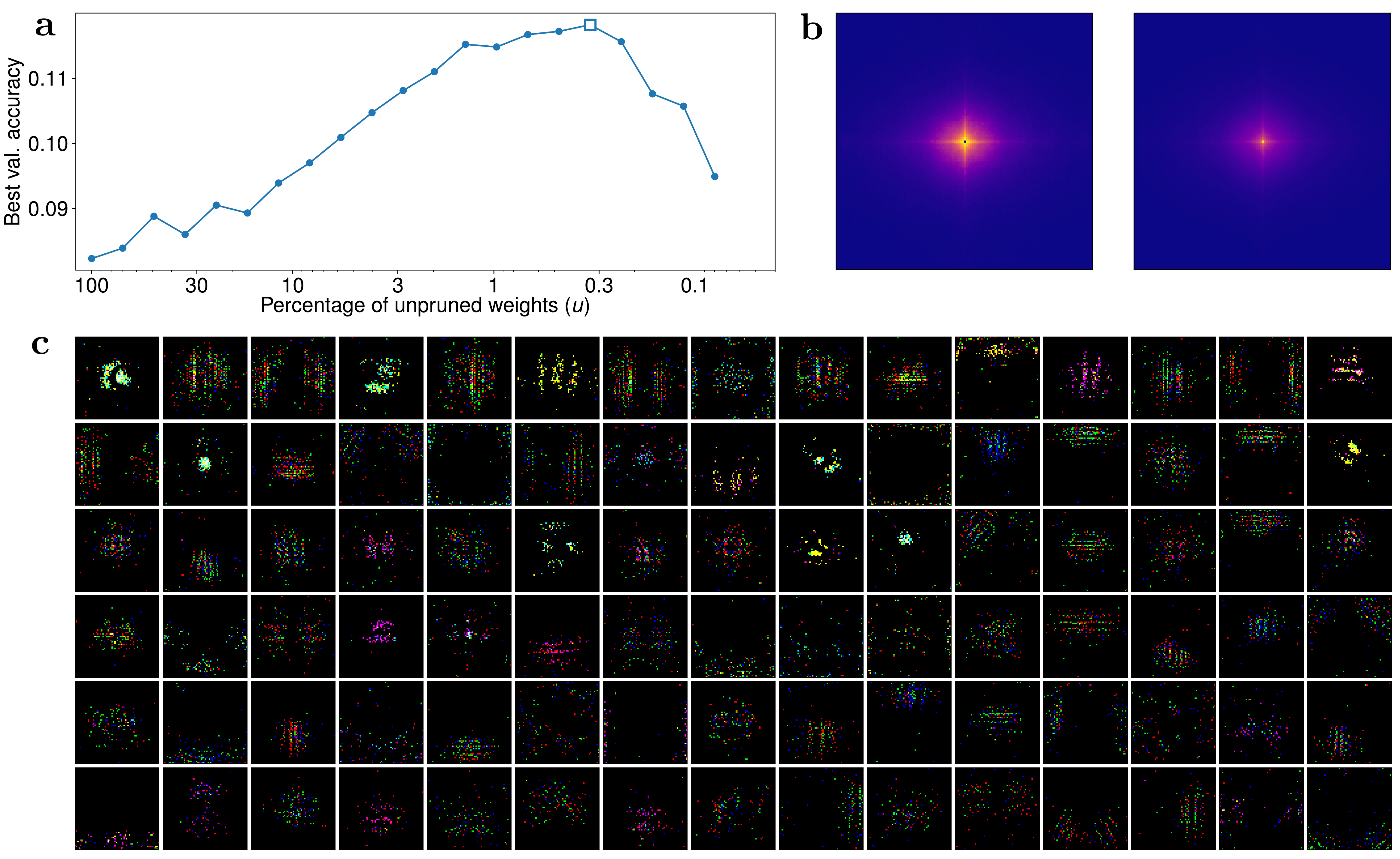}
  \caption{
    \textbf{a:} Best validation accuracy as a function of $u$ for ImageNet64 dataset.
    \textbf{b:} $S^{\rm sc}$ and $S^{\rm dc}$ (see Fig.~\ref{fig:2} for details) for the best iteration at $u\sim0.3\%$ (square in panel \textbf{a}).
    \textbf{c:} 90 most connected layer 1 masks for the same iteration.
  }\label{fig:IN64}
\end{figure}

We train a network on the $64\times64$ pixels, higher resolution ImageNet downsampling from the same source: ImageNet64~\cite{ImageNet32}.
We use exactly the same parameters of our main experiment, and obtain only slightly lower accuracy and a similar IMP curve as the lower resolution version, as shown in Fig.~\ref{fig:IN64}\textbf{a}.
We analyze the best iteration at $u\sim0.3$: locality is still preserved as shown in panel \textbf{b}, with the main axes being preferred even more sharply.
The 90 most connected layer 1 masks (panel \textbf{c}) show similar features to the ones found elsewhere, with even clearer preference for sparse sampling and separated lines along the main orientations.

\subsection{Translated images}\label{SMss:transl}
\begin{figure}[!htb]
  \centering
  \includegraphics[width=\columnwidth]{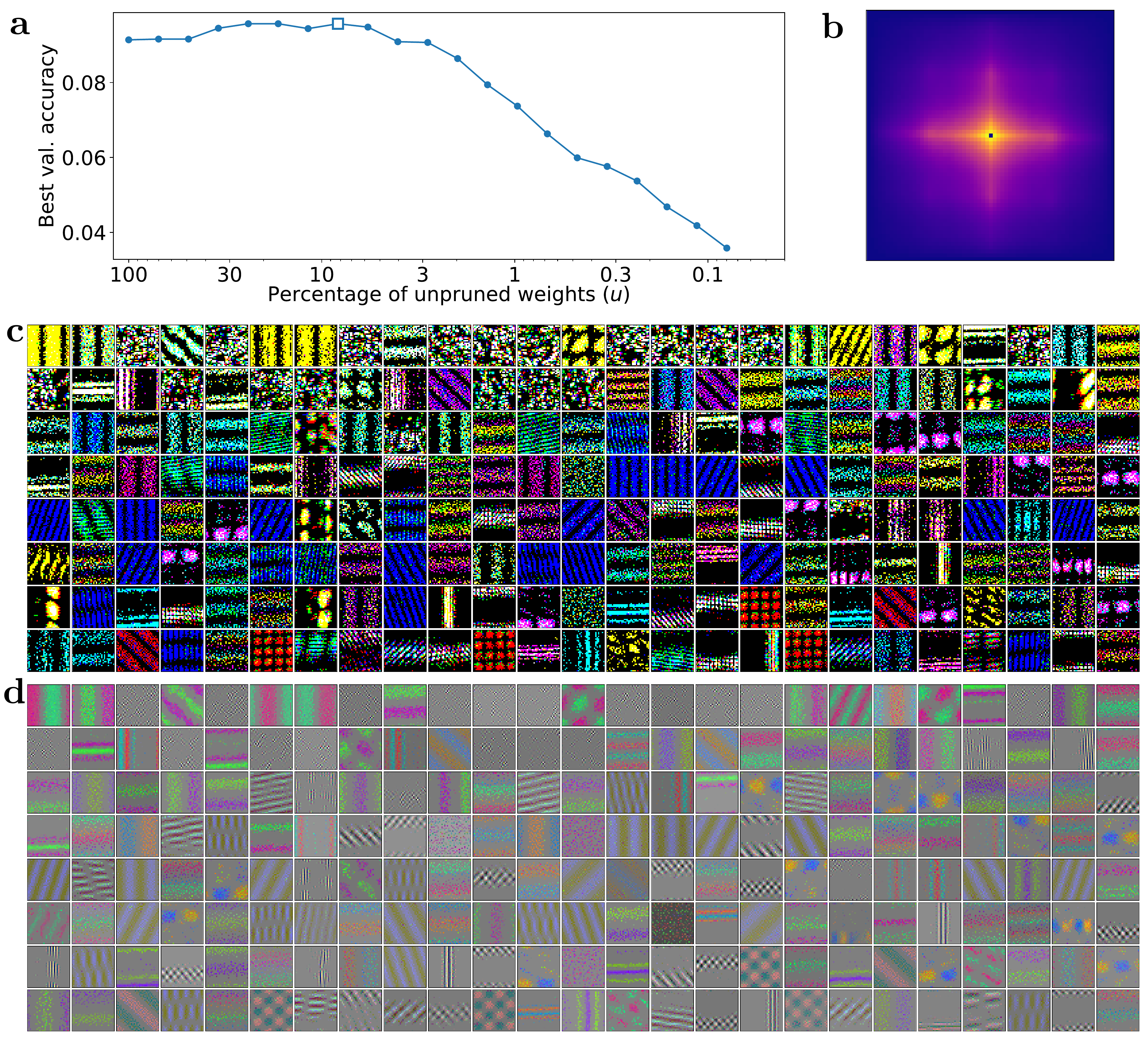}
  \caption{
    \textbf{a:} Best validation accuracy as a function of $u$ for randomly translated dataset.
    \textbf{b:} $S^{\rm sc}$ (see Fig.~\ref{fig:2} for details) for the best iteration at $u\sim8\%$ (square in panel \textbf{a}).
    \textbf{c:} 200 most connected layer 1 masks for the same iteration.
    \textbf{d:} The same masks, multiplied by the relative weights (see Fig.~\ref{fig:3} for details).
  }\label{fig:transl}
\end{figure}

For this experiment, we enforce complete translational invariance by translating each image vertically and horizontally by a random amount (in $[0,31]$) modified every time a mini-batch is created, and wrapping the image in both axes (periodic boundary conditions). 
This clearly causes the appearance of unrealistic boundaries in the image, but achieves the goal of this experiment or rendering each pixel is equivalent, on average.
We validate this network on the original, untranslated images, achieving lower accuracy then the original network, as might be expected since the training dataset is much larger and not matching the validation dataset (we also do not re-optimize the training parameters, and just run long trainings of 500k steps).
The IMP validation curve in Fig.~\ref{fig:transl}\textbf{a} shows a peak at lower sparsity ($u\sim8\%$), consistent with the modified dataset.
Analyzing this best iteration, the $S^{\rm sc}$ map in panel \textbf{b} indicates local connectivity, with a stronger bias towards the horizontal and vertical directions.
The large variety of maps in panel \textbf{c} (and their weightd counterparts in panel \textbf{d}) show surviving features in the shape of stripes and patches patterns with different orientations and periodicity.
Despite the synthetic nature of this dataset, it is interesting to observe these patterns emerge from a translationally invariant dataset. Moreover, the combination of these features with the spatial focus of the original dataset (also visible in Fig.~\ref{fig:ref2}) helps explain the emergence of Gabor-like filters.

\section{External resources}\label{SMs:ext}
A repository with all the code required to reproduce the results of this paper is available at \url{https://github.com/phiandark/SiftingFeatures}. The main code is written in Python and based on tensorflow (1.xx). Input files are provided to reproduce our experiments and a jupyter notebook is available to post-process the data and recreate the figures of this work.

The same repository hosts a text file with the attribution of each ImageNet category to one of the 10 macro-class in our semantic clustering experiment.

Videos are also available in the same location, showing the evolution of the most connected masks and their weighted versions during IMP for the main task of this work.

\end{document}